\pdfoutput=1
\documentclass[11pt]{article}

\usepackage[]{emnlp2021}

\usepackage{times}
\usepackage{latexsym}
\usepackage{hyperref}

\usepackage[T1]{fontenc}
\usepackage[utf8]{inputenc}
\usepackage{graphicx}
\usepackage{amsmath}
\usepackage{booktabs}
\usepackage{microtype}

\title{Do Long-Range Language Models Actually Use Long-Range Context?}

\author{
Simeng Sun$^1$ \hspace{3mm} Kalpesh Krishna$^1$ \hspace{3mm} Andrew Mattarella-Micke$^2$ \hspace{3mm} Mohit Iyyer$^1$ \\
University of Massachusetts Amherst$^1$  Intuit AI$^2$\\
  {\tt \{simengsun,kalpesh,miyyer\}@cs.umass.edu} \\
  {\tt andrew\_mattarella-micke@intuit.com} 
  }

\begin{document}
\maketitle

\begin{abstract}
Language models are generally trained on short, truncated input sequences, which limits their ability to use discourse-level information present in long-range context to improve their predictions. Recent efforts to improve the efficiency of self-attention have led to a proliferation of long-range Transformer language models, which can process much longer sequences than models of the past. However, the ways in which such models take advantage of the long-range context remain unclear. In this paper, we perform a fine-grained analysis of 
%the \emph{Routing Transformer}~\citep{roy2020efficient},
two long-range Transformer language models (including the \emph{Routing Transformer}, which achieves state-of-the-art perplexity on the PG-19 long-sequence LM benchmark dataset)
that accept input sequences of up to 8K tokens.
Our results reveal that providing long-range context (i.e., beyond the previous 2K tokens) to these models only improves their predictions on a small set of tokens (e.g., those that can be copied from the distant context) and does not help at all for sentence-level prediction tasks.
%Broadly, our results reveal that the distant context (i.e., beyond the previous 2K tokens) is almost never useful for token or sentence-level prediction tasks. 
%Furthermore, the few improvements come from \emph{copying} tokens that occur in the distant context, regardless of the context in which these tokens appear (the model is insensitive to word order in the distant context). 
Finally, we discover that PG-19 contains a variety of different document types and domains, and that long-range context helps most for literary novels (as opposed to textbooks or magazines). 
\end{abstract}
\section{Introduction}

Understanding long documents requires modeling various discourse-level phenomena, including anaphora~\citep{hobbs1979coherence,grosz-etal-1995-centering}, argument structure~\citep{grimshaw1990argument}, narrative scripts and trajectories~\citep{schank1977scripts,labov1997narrative}, and causal links between concepts~\citep{mooney1985learning}. Unfortunately, most language models (LMs) are trained to predict the next word given only a small window of local context, which prevents them from using long-range discourse structure to improve their predictions. Many research efforts over the years have attempted to address this issue: for example, ~\citet{rosenfeld1996maximum} incorporated statistics from distant tokens to improve $n$-gram models, while~\citet{ji2015document} added document-level context to neural LMs.

More recently, the Transformer LM~\citep{vaswani2017attention}, which forms the backbone of state-of-the-art NLP systems~\citep{devlin2019bert,brown2020language}, has become the focus of numerous efforts to process longer input sequences. Transformer LMs are constrained by the inefficiency of the self-attention mechanism, whose complexity scales quadratically with the input sequence length. As such, more efficient methods based on sparse attention~\citep{correia-etal-2019-adaptively} and cached memory~\cite{Rae2020Compressive} have been proposed to increase the maximum sequence length, which has progressed from GPT-2's 1024 tokens~\citep{radford2019language} to 4096 tokens~\cite{zaheer2020big} and finally 8192 tokens~\cite[][\emph{Routing Transformer}]{roy2020efficient}. When evaluated on the PG-19 benchmark dataset~\citep{Rae2020Compressive}, which contains long documents in the public domain, these ``long-range'' Transformer LMs also reach lower perplexities than baseline models on held-out data.

How do ``long-range'' Transformer LMs make use of the long-range context? Do they actually encode important discourse information to improve their predictions? In this paper, we conduct a series of fine-grained analysis experiments to answer these questions, several of which are inspired by the context analysis of LSTM LMs conducted by~\citet{khandelwal-etal-2018-sharp}. We focus specifically on analyzing the behavior of the state-of-the-art Routing Transformer and a simpler baseline model in the presence of various perturbations (e.g., word shuffling, random document replacement), and look closely at how different types of tokens  are affected. 
Our results show that:
\begin{itemize}

% (2K)
\item Providing long-range context (i.e., further than 2K tokens away) to these models has \emph{negligible} impact on the perplexity of tokens near the end of a sequence in aggregate. However, a fine-grained analysis reveals that it does help a small set of tokens (tokens within subword clusters and those that can only be copied from the distant context), as well as particular types of books (fictional and continuous).
%but is not useful for sequence-level prediction tasks.

% (perturbation)
\item Despite the aforementioned improvements on a fraction of tokens, significantly perturbing the long-term context with word shuffling and random replacement has no notable impact on perplexity overall, suggesting that the evaluated models encode long-range context superficially at best.

\item Long-range context is not used for sequence-level prediction tasks that move outside the teacher-forced setting of the previous experiments.

% (book types)
% \item After manually annotating the entire validation set of PG-19 by book genre and continuity, we find that long-range context is more beneficial for fictional and continuous books.

\end{itemize}

While modern LMs can process much longer input sequences than those of the past, we conclude that they do not exploit the information available in the long-range context. We recommend that future research on long-range LMs includes analysis experiments such as those in our work to shed light on how and when they are using the distant context.

\section{Background \& Setup}

In this section, we first provide an overview of the long-range language models analyzed in this work (Local Transformer, Routing Transformer). Next, we describe the experimental setup used in the remainder of the paper to measure the impact of the long-term context.

\subsection{Long-range Language Modeling} 
Given preceding context tokens $w_{<i}$ (the \emph{prefix}), a language model (LM) computes the probability distribution of the next token $p(w_i \mid w_{<i})$. LMs are commonly evaluated using perplexity, which is the exponentiated negative log likelihood of a held-out corpus:
$$ ppl = \exp \bigg(-\frac{1}{N}\sum_{i=1}^N \log p(w_i \mid w_{<i}) \bigg)$$

Modern LMs are most often implemented with \emph{Transformers}~\citep{vaswani2017attention}, which compute vector representations for every token in the prefix at multiple layers and combine them together using self-attention. Since self-attention requires scoring every token in the sequence against every other token, it scales quadratically in both compute time and memory usage, which limits its application to very long sequences.

\paragraph{Local Transformer} A simple way to improve the efficiency of self-attention blocks is to constrain the attention at each layer to a local window of the previous $k$ tokens. Such Transformers, which we refer to as \emph{Local Transformers}, can be feasibly scaled up to large input sequence lengths. The receptive field of a Local Transformer scales linearly with the number of layers, as the $l^{th}$ layer of this model can access the previous $k \times l$ tokens~\cite{luong-etal-2015-effective, child2019generating,sukhbaatar-etal-2019-adaptive}.
% \micomment{i think there are other citations since hten that use this in transformer variants? cant remember which, maybe RT paper describes them}

\paragraph{Routing Transformer} The Routing Transformer~\citep[][RT]{roy2020efficient} takes a more intelligent approach to scaling self-attention. Specifically, the RT assigns keys and queries in self-attention to $k$ clusters, the centroids of which are learned during training. A routing attention strategy computes attention $A$ only over the queries $Q_i$ and keys $K_j$ that belong to the same cluster $\mu(Q_i)$ (i.e., those whose centroid is closest to query $Q_i$).
% let $A$ be the attention matrix, where $A_{i,<i}$ represents the attention weights of token $w_i$ on previous tokens. 

\[
    X_i = \sum_{\substack{j:K_j\in\mu(Q_i),\\ j<i}} A_{ij}V_j
\]

% Specifically, token representations are first clustered using the $k$-means algorithm, and only tokens within the same cluster are permitted to attend to each other. 
In contrast to the position-based local attention, this clustering-based attention takes the \emph{content} of the token representations into account. 
This sparse self-attention mechanism reduces the complexity from $O(N^2)$ to $O(N^{1.5})$ and has led to state-of-the-art results on tasks such as long-form question answering~\citep{krishna2021hurdles}.
% \micomment{mention some applications of RT, eg kalpeshs lfqa paper, to further make the point that this is a practically useful model}

\subsection{Experimental setup}
While the previously-described models can be trained with longer inputs than standard Transformers, it remains unclear how they make use of the additional context tokens to improve their predictions of the next word. To shed light on the behavior of long-range Transformer LMs, we perform a series of experiments in which we manipulate the input sequence (both length and content). For token-level experiments (\S~\ref{section:context-size}, \S~\ref{section:context-perturb}), we only evaluate the perplexity of $k$ tokens near the end\footnote{An artifact exhibited by RT causes tokens at the very end of a sequence to have much higher losses than the others; this phenomena does not exist with LT. After correspondence with the RT authors, we decide to exclude the very last 40 tokens (whose losses are affected) from all of our evaluations. More details about this artifact can be found in the Appendix.} of an $N$ token-long input sequence ($k \ll N$) to focus on the effect of long-range context.\footnote{Language models are normally evaluated on non-overlapping sequences, with the begin and end sequence tokens receiving different amount of context. In our setting, all target tokens have roughly the same amount of context.}
% \micomment{add footnote contrasting this with the standard way that RT evals perplexity}

\paragraph{Dataset:}
We conduct all of our analyses on the validation set of the PG-19 dataset~\citep{Rae2020Compressive}. This dataset contains $\sim$29K books from Project Gutenberg repository published before 1919 and was constructed specifically to evaluate long-range LMs (average document length $\sim$69K tokens). The validation set contains 50 books\footnote{We observe a significant gap of $\sim$10 perplexity between the PG-19 test and validation sets and discover that this gap is largely due to the presence of an annotated edition of \textit{The Canterbury Tales and Other Poems}. This book intertwines line-by-line annotations with the main text, which causes the preprocessed version in the dataset to be unreadable. We remove this book in all of our experiments, which decreases the test/validation perplexity gap to $\sim$3.} and $\sim$3 million tokens in total.
% \micomment{need to add more details about this dataset. what kind of texts are in it, how big is it , how many books in val set, avg doc length, etc}. The validation set contains 49 books with average book length 56K...
% 2725337
% 55619.12244897959
Evaluating every token in the validation set with large prefix sizes (e.g. 8K tokens) is computationally infeasible.\footnote{On one RTX8000 GPU, it takes around 104h to evaluate the entire PG-19 validation set with sequence length 8K and target sequence 10.} Thus, we set the number of target tokens per context $k=10$ and sample a subset of 220K validation tokens for our experiments, which is the same data size used in the LM analysis experiments of~\citet{khandelwal-etal-2018-sharp}. Evaluating RT in this way yields slightly better perplexity on the validation set of PG-19 than using the evaluation setting in the original RT paper (35.2 vs. 36.3).
% \micomment{add something saying evaluating the model this way yields similar test perplexity to that reported in the paper and put the numbers} 
We ensure that the number of tokens sampled from each validation book is proportional to the length of that book. 

% 490710140
\paragraph{Models:}
As training long-range Transformer LMs is also infeasible without immense computational resources, we use publicly-available pretrained checkpoints for all of our experiments. The Routing Transformer (RT) checkpoint contains 490M parameters and processes sequences up to 8192 tokens long, achieving 33.2 perplexity on the PG-19 test set. The released checkpoint, which was trained on 128 TPUv3 cores for several weeks, has a subword vocabulary of $\sim$ 98K types,\footnote{We follow the RT paper~\cite{roy2020efficient} by scaling the loss by 1.248 before computing the perplexity in order to match the word-level perplexity reported by \citet{Rae2020Compressive}.} along with 8 attention heads in each of its 22 layers.
% \micomment{how many heads per layer, how many params, mention test ppl, mention max seq length, vocabulary size + tokenization}
The top two RT layers include content-based clustering attention while the remaining are composed entirely of local attention.
% The released model was trained on 128 TPUv3 cores with 1.4 sec/step,
% for 1919h 
% \footnote{\sscomment{Estimated through step/sec reported in the paper and training steps from the checkpoint.}},
% which is infeasible to train from scratch by ourselves.  
% \micomment{add how much training time/how many TPUs used to make it clear that its not feasible to train these things ourselves. also add something about how we verify that the clustering heads incorporate info across the long-range context} 

\begin{figure}
    \centering
        \includegraphics[width=0.35\textwidth]{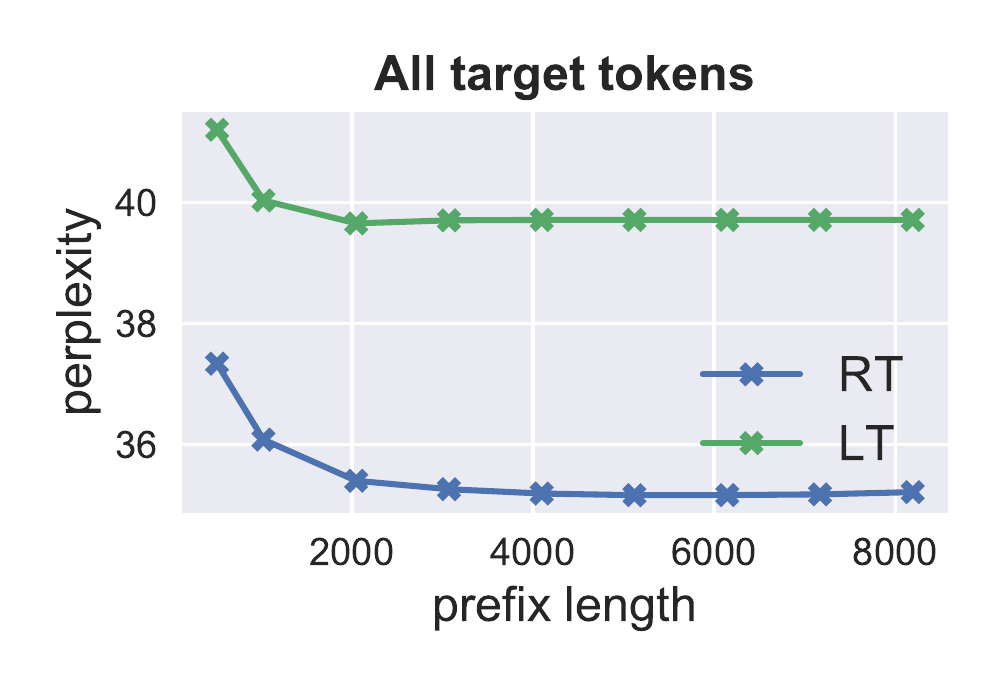}
    \caption{The perplexity of all target tokens plateaus after 2K prefix tokens for both Routing Transformer (\textbf{RT}) and Local Transformer (\textbf{LT}), showing the negligible overall impact of the long-range context.
    }
    \label{fig:length-overall}
\end{figure}

Our Local Transformer (LT) is derived from the same checkpoint as RT (and thus has identical model size), except that all clustering heads are replaced with local attention heads. It achieves slightly better perplexity on the PG-19 test set (38.3 vs. 39.3) compared to the LT model trained from scratch by~\citet{roy2020efficient}, possibly because the local attention heads learn a better representation of the weight matrices by using the information from the clustering heads.\footnote{The original fully trained LT checkpoint was not made publicly available before the EMNLP submission deadline.}\footnote{The RT authors released a new LT checkpoint during the review period of this paper. We also evaluate this newly-released LT checkpoint and include the results in the Appendix~\ref{section:appendix-new-lt}. Both the RT and the LT checkpoints can be found at \url{https://github.com/google-research/google-research/tree/master/routing_transformer} }  The attention heads in this model attend to the previous 256 tokens, which results in an effective receptive field of $\sim$ 5K tokens.\footnote{A preliminary experiment verified that the clustering heads in the RT do attend to the long-range context, beyond 5K tokens, demonstrating that it is at least theoretically incorporating more context than the LT.}
% \sscomment{todo: reason for why not using other models.}
While we would have also liked to analyze other long-range LMs such as the Compressive Transformer~\cite{Rae2020Compressive} and Longformer~\cite{beltagy2020longformer}, these models do not have publicly-available PG-19 checkpoints; additionally, they differ from RT and LT in model size, which makes it hard to perform controlled experiments.

\section{The effect of longer context} \label{section:context-size}

% Previous analysis by~\citet{khandelwal-etal-2018-sharp} shows the LSTM LM has effective context size of around 200. Recent advances in long-range LMs enable the Transformer to accept much longer context size, however, it's unclear how these Transformer LMs take advantage of the distance context. 
How does the size of the prefix affect the perplexity of  long-range Transformer LMs? In this section, we evaluate our RT and LT checkpoints on the PG-19 validation set with varied prefix length. We discover that although these models are theoretically able to encode long sequences, increasing the prefix length beyond 2K tokens does not bring discernible improvements in aggregate. However, we do identify small subsets of tokens that benefit from long-range context. Additionally, we find that these models take advantage of long-range context to different degrees on different types of books (e.g., continuous fictional narratives vs. discontinuous magazine articles).

\paragraph{Validation perplexity does not improve when the prefix length grows beyond 2K tokens: }
As shown in  Figure~\ref{fig:length-overall}, RT perplexity plateaus when evaluated with prefixes longer than 2K.\footnote{As a point of comparison, the perplexity of the much smaller 
LSTM language models evaluated by ~\citet{khandelwal-etal-2018-sharp} plateaus after 200 words. Additionally,~\citet{press2020shortformer} discover that the perplexity flattens after 1K for a smaller standard Transformer LM.
} In contrast, relative to RT, the perplexity curve for the more primitive LT starts flattening earlier at around 1K tokens (note that its effective context size is only 5K). We conclude that while RT's clustering heads take better advantage of global context than LT, the long-range context beyond 2K tokens is not helping overall. Surprisingly, the perplexity gap between RT and LT is relatively consistent regardless of the prefix length, which indicates that much of RT's gains do not come from its increased ability to leverage long-range context but rather from better modeling of \emph{local context}. 

\begin{figure}
    \centering
    \includegraphics[width=0.45\textwidth]{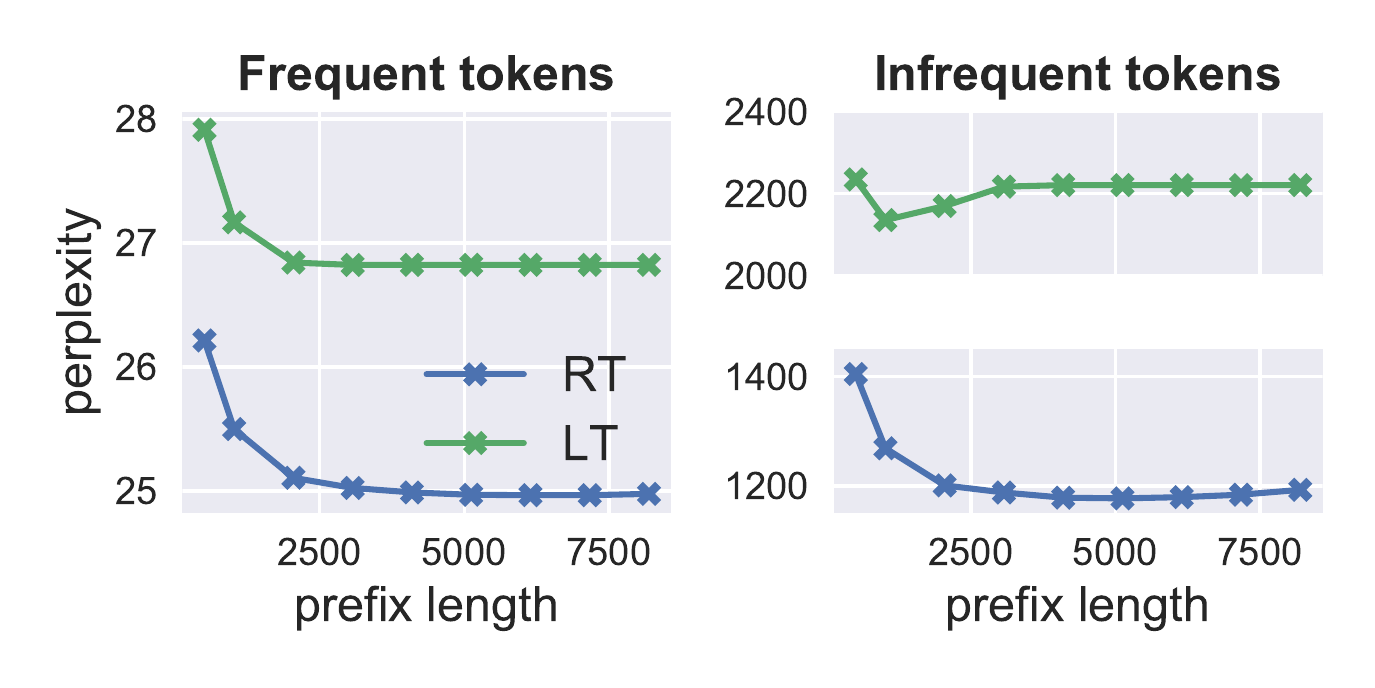}
    \caption{RT perplexity of infrequent tokens continues to decrease until prefix length is 5K.
    % beyond 2K prefix tokens.
    }
    \label{fig:length-and-freq-vs-length}
\end{figure}
\begin{figure}
    \centering
    \includegraphics[width=0.45\textwidth]{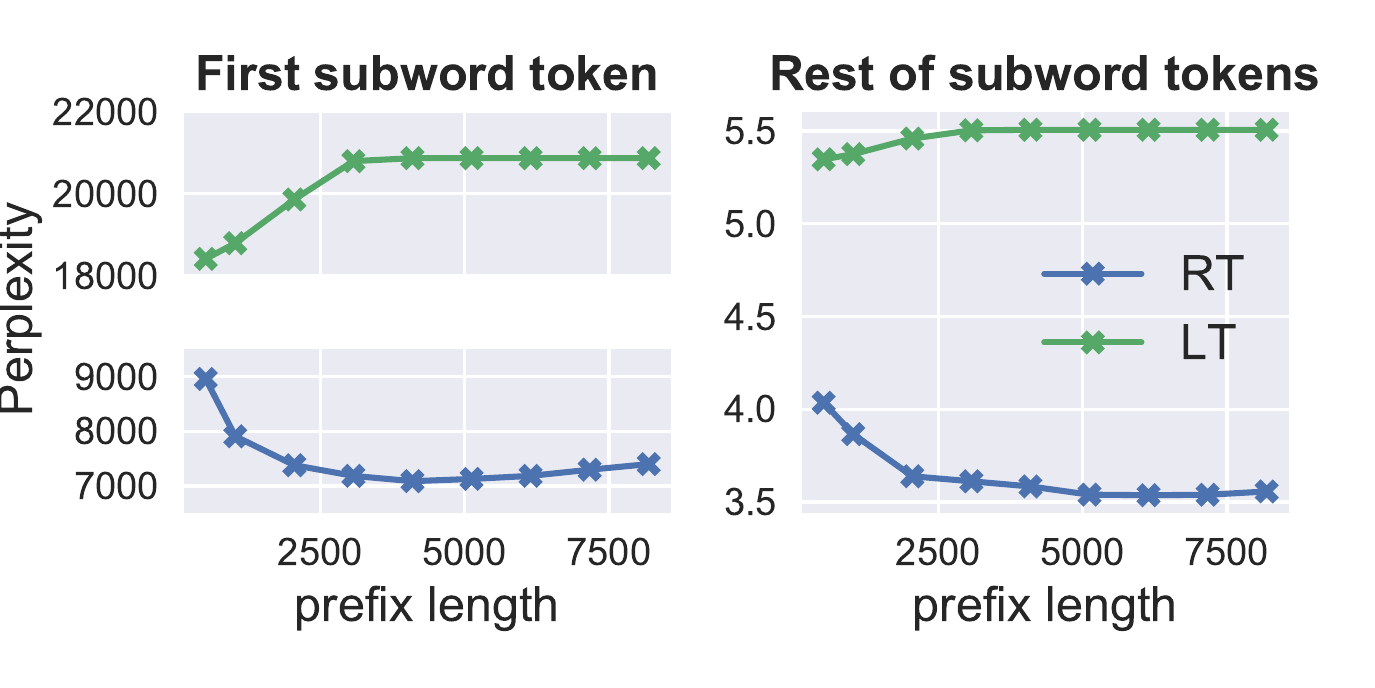}
    \caption{RT perplexity of tokens inside subword clusters continues to decrease until a prefix length of 5K.}
    \label{fig:subwords-vs-length}
\end{figure}

\paragraph{Infrequent tokens can benefit from increased prefix length:} 
While the overall perplexity does not improve with increasing prefix length, we do observe different behavior when filtering the target tokens by frequency, as shown in Figure~\ref{fig:length-and-freq-vs-length}. We define \emph{frequent} tokens to be the top 10\% more frequently-occurring tokens in the subword vocabulary of PG-19 while the rest of tokens are classified as \emph{infrequent}.\footnote{Around 20K tokens in our target token set are classified as infrequent, which amounts to 9\% of all target tokens.}
% \micomment{how many frequent and infrequent tokens are used to make the plots? need to be more specific. also is the frequency measured at word or subword level?} 
While adding long-range context does not improve either model's predictions of frequent tokens, the RT's perplexity of infrequent tokens decreases from $\sim$ 1200 with a prefix length of 2K to $1180$ with prefix length of 5K. However, we do observe that infrequent token perplexity increases back to 1200 as the input is further extended, suggesting that the additional context perhaps confounds the model. Meanwhile, the LT is significantly worse than RT on infrequent token prediction, and its perplexity increases as the prefix length is increased.\footnote{This is likely an artifact due to the elimination of routing attention heads from the RT checkpoint, since the LT trained from scratch does not exhibit such behavior. More details are included in Appendix~\ref{section:appendix-new-lt}.}

\begin{figure}
    \centering
    \includegraphics[width=0.45\textwidth]{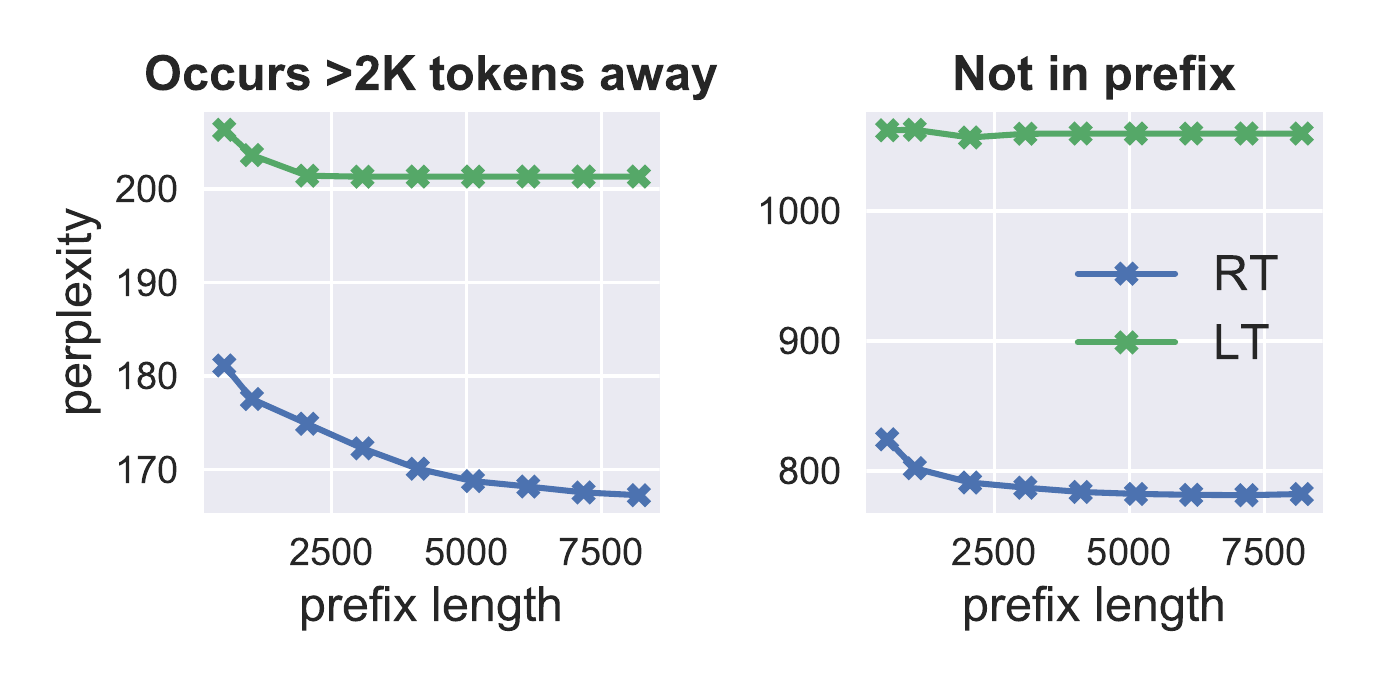}
    \caption{ RT perplexity of target tokens whose last appearance is more than 2K tokens away in the prefix keeps decreasing.
    }
    \label{fig:copy-ncopy-vs-length}
\end{figure}

\begin{figure}
    \centering
     \includegraphics[width=0.46\textwidth]{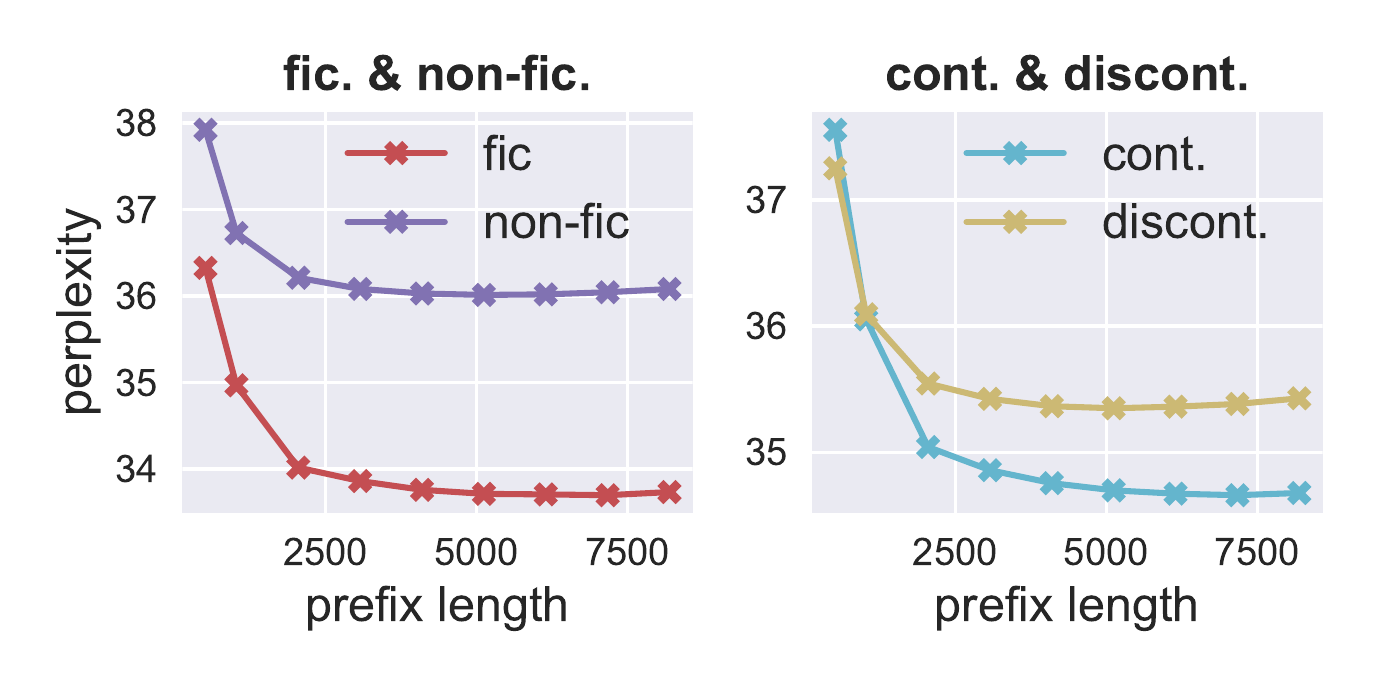}
    \caption{ RT takes better advantage of context beyond 2K tokens for fictional and continuous books than non-fictional and discontinuous books, respectively.
    }
    \label{fig:distant-by-book-types}
\end{figure}

\paragraph{Tokens inside a subword cluster benefit from longer contexts:} One issue with the previous experiment is that the frequency categorization was computed at a subword level and so may not exactly correspond to word frequency, especially for infrequent words (e.g., entities) that are split into multiple subwords. We therefore do a follow-up experiment by isolating all words that are split into multiple subwords, and then examining perplexity of these tokens as a function of their position in the subword cluster. For example, the word ``Trocadero'' is separated into three subword tokens ``Tro'', ``cade'', and ``ro''. 
We specifically distinguish between the \emph{first} subword in the cluster (``Tro'') from the \emph{rest} of the subwords (``cade'' and ``ro'') in the plots shown in Figure~\ref{fig:subwords-vs-length}. The perplexities are computed over 4.1K \emph{first} and 5.1K \emph{rest} subword tokens.
% \micomment{write how many first vs rest subwords are included in the plots} 
The \emph{first} subword category exhibits the same curve shape as those for infrequent tokens for both models, although the magnitude of the perplexities is far higher.  The rest of the subwords are far easier for both models to predict, but the RT perplexity curve shows some positive impact from the long-range context until a prefix size of 5K tokens.

\paragraph{Routing Transformers are able to copy tokens that occur in the long-range context:} 
Target subword tokens that can be copied from somewhere in the prefix form another interesting group to analyze.\footnote{Note that there is some overlap between the token categories we have analyzed so far. We verify in the Appendix~\ref{section:token-overlaps} Table~\ref{tab:overlap-ratio} that the overlap between these categories is not significant enough to confound the results.} While this is commonplace for frequent words (e.g., determiners, pronouns), it also occurs for entities and rare words (e.g., character names in a novel); sometimes, a word can occur several thousand tokens after its last occurrence. We focus on the latter category of tokens, specifically using a prefix length of 2K tokens as a cutoff to distinguish local and long-range context. 
Perplexities are computed over 22k tokens which occur last time more than 2K tokens away, and 36K tokens that never appear in the prefix.
% {write how many pre2048 vs nocopy tokens are included in the plots} 
In particular, the left plot in Figure~\ref{fig:copy-ncopy-vs-length} shows the perplexity of tokens that cannot be found in the previous 2K tokens, but occur somewhere in the long-range context (2K to 8K tokens away). While the LT curve for such tokens plateaus after 2K tokens, indicating that LT cannot take advantage of repeated words in the distant context, the RT curve steadily decreases until 8K tokens. The right plot, which shows the subset of target tokens which do not occur anywhere in the short or long-term context, decreases until about 5K tokens and then plateaus. Overall, these results show that long-range context is helpful for tokens that appear even several thousands tokens away. 

\begin{figure}
    \centering
     \includegraphics[width=0.46\textwidth]{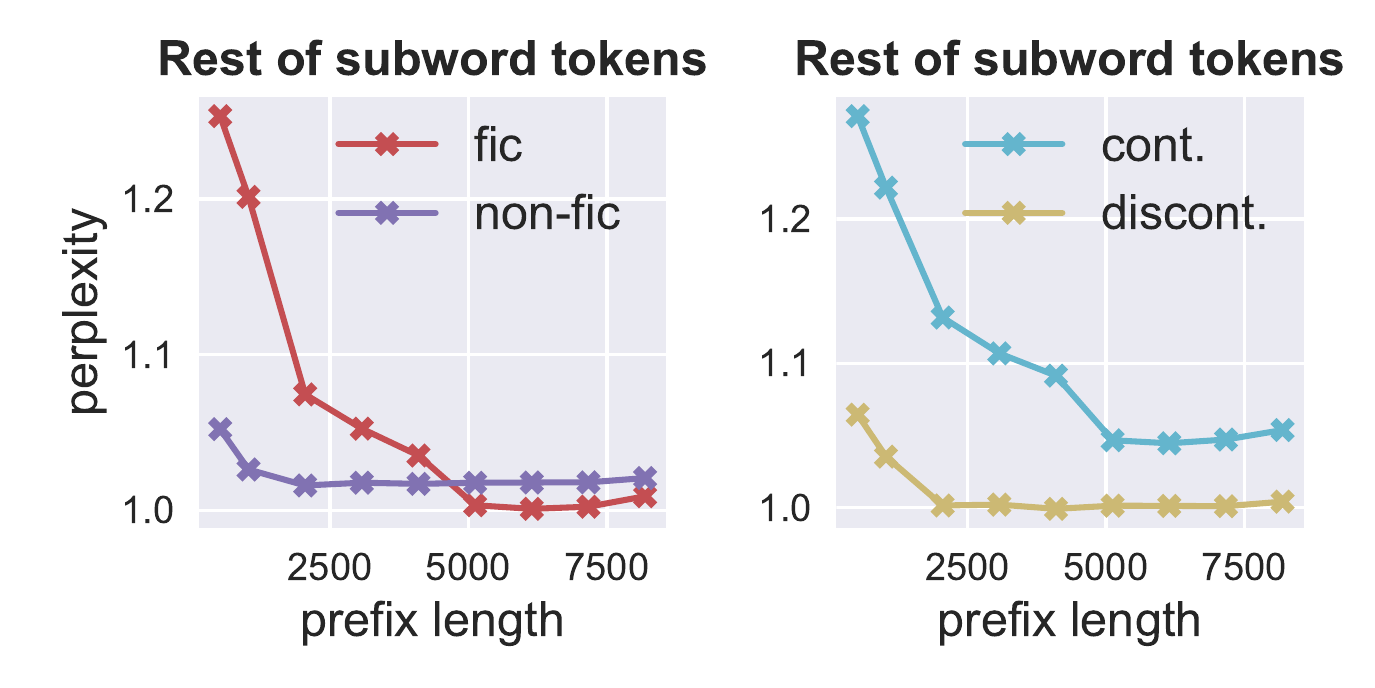}
    \caption{The majority of improvements on tokens inside subword clusters (i.e., excluding the first token in subword clusters) from prefixes longer than 2K tokens comes from fictional and continuous books.
    }
    \label{fig:subword-by-genre}
\end{figure}

\paragraph{Following patterns in the long-range context:} Besides the token categories examined above, we also qualitatively look at some examples that are too infrequent to analyze at scale. Interestingly, we observe some simple patterns (slightly more complex than copying) that the RT model picks up on. Specifically, it learns to increment chapter numbers even if the previous chapter title appears more than 2K tokens away: for example, when predicting ``Chapter V'' in the validation book \emph{Keith of the Border}, modifying the previous chapter title ``Chapter IV'', which occurs 2300 tokens away, to ``Chapter V'' causes the loss of the predicted token ``V'' to increase by over 10.

\begin{figure}
    \centering
    \begin{minipage}{0.35\textwidth}
    \includegraphics[width=\textwidth]{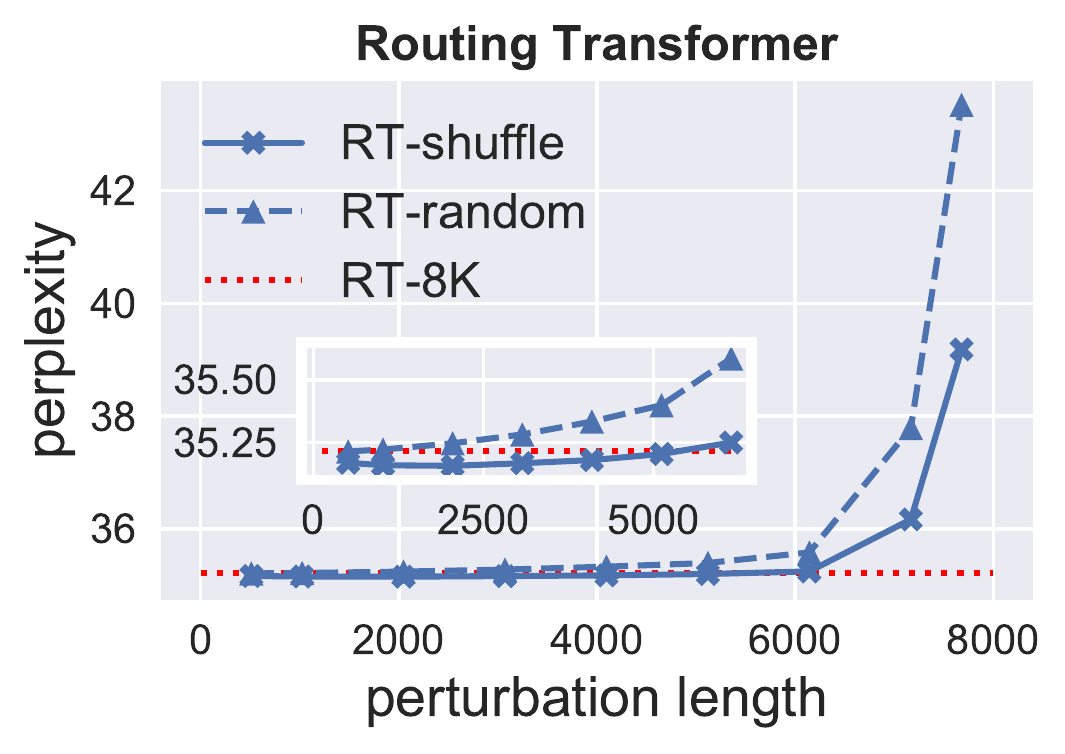}
    \end{minipage}
    \caption{
    Perturbing up to 6K prefix tokens does not notably affect RT's overall perplexity. 
    The corresponding plot for LT is included in Appendix~\ref{section:appendix-context-perturb}.}
    \label{fig:perturb-overall-vs-length}
\end{figure}

\begin{figure}
    \centering
    \includegraphics[width=0.45\textwidth]{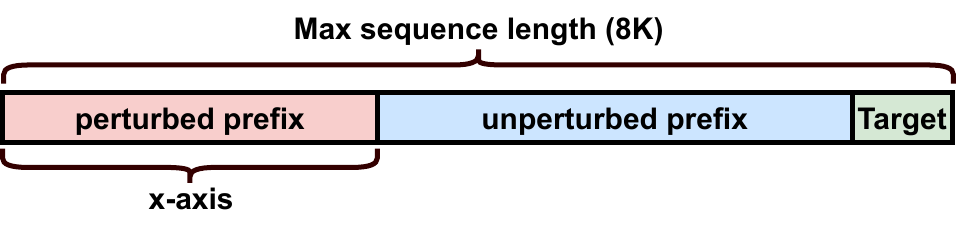}
    \caption{Generic illustration of the x-axis in all the perturbation analysis.}
    \label{fig:x-axis}
\end{figure}

\paragraph{The impact of book type on the benefits of long-range context:} \label{section:book-benefit-from-lrc}
PG-19 contains a diverse array of topics, genres, and formats, not all of which equally benefit from long-range context modeling. For example, while continuous narratives (e.g., novels) certainly build up many high-level discourse structures over a long sequence of tokens, discontinuous text like magazines, textbooks, or short story collections may require primarily local modeling. To better understand the effect of the type of book on long-range LM perplexity, we annotate every book in PG-19's validation set as either \emph{fiction} or \emph{non-fiction} and \emph{continuous}\footnote{We consider books with related but distinct sections (such as textbooks) to be discontinuous in our annotation.} or \emph{discontinuous}.\footnote{We also annotate whether the work has been written by the same author or various authors, which is presented in the Appendix.} Out of 49 books we annotated, 30 are non-fiction,\footnote{Some magazines contain short stories or poems interspersed with news articles and essays; we count these as non-fiction in our analysis.} 31 are discontinuous, and 25 books are both non-fiction and discontinuous.

We observe in Figure ~\ref{fig:distant-by-book-types} that the RT model takes better advantage of long-range context for fictional and continuous books, as the perplexity for these books plateaus at around 5K tokens.
% \textcolor{cyan}{For the previously described subset of tokens where adding more context keeps improving the perplexity, we also evaluate those tokens broken down by book types.
Figure~\ref{fig:subword-by-genre} shows fictional and continuous books exploit better the long-range context while predicting tokens within subword clusters. Overall, we find the improvement stems largely from continuous and fictional books; more details are included in Appendix~\ref{section:appendix-book-type}.

\section{The perturbation of long-range context} \label{section:context-perturb}

The experiments in the previous section show that incorporating long-range context (further than 2K tokens away from the target) yields only marginal improvements to the overall perplexities of RT and LT. However, the long-range context does have a notable positive impact on a subset of tokens and book types. Do these improvements persist in the presence of severe perturbations to the distant context? If so, this would indicate that they are not encoding any complex discourse structure in the context but rather relying on surface information (e.g., token presence) to make better predictions. In this section, we perform a perturbation analysis to quantitatively measure the robustness of the state-of-the-art RT model.\footnote{Figure~\ref{fig:perturb-overall-LT} in the Appendix shows that the Local Transformer never uses context beyond 3K. Due to this limitation, we only present results on RT for in this section.}

Formally, assume we are given a prefix sequence $P=(w_0, w_1, \dots, w_n)$ with which we want to predict target sequence $(w_{n+1}, w_{n+2}, \dots, w_{n+k})$. We apply a perturbation $\rho$ to the first $m$ tokens of the prefix ($w_{0:m}$) to obtain the perturbed prefix 
\begin{equation*}
\Tilde{P}=(\rho(w_0,\dots,w_m), w_{m+1}, \dots, w_n).
\end{equation*}

We define the following three perturbation operations for $\rho$ and report results averaged over five runs for each of them.
\begin{itemize}
    \item \textbf{Sequence shuffling}: Tokens within the perturbed window $w_{0:m}$ are shuffled across the entire window (i.e., sentence boundaries are not respected).
    \item \textbf{Random sequence replacement}: $w_{0:m}$ is replaced with a random sequence from another validation book that is $m$ tokens long.
    \item \textbf{Specific token drop}: Specific tokens within $w_{0:m}$ (e.g., those that occur in the target) are dropped and replaced with the padding token.
    % \item \textbf{Specific token replacement}: Specific tokens within $w_{0:k}$ are replaced with a randomly-sampled token from the vocabulary.
\end{itemize}

\begin{figure}
\centering
    \begin{minipage}{0.225\textwidth}
    \includegraphics[width=\textwidth]{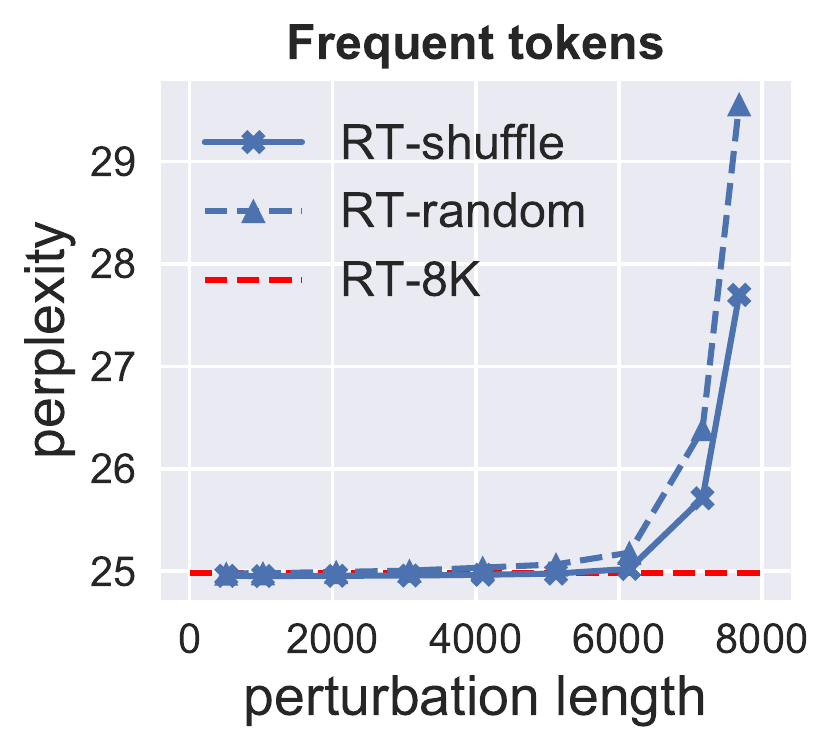}
    \end{minipage}
    \begin{minipage}{0.22\textwidth}
    \includegraphics[width=\textwidth]{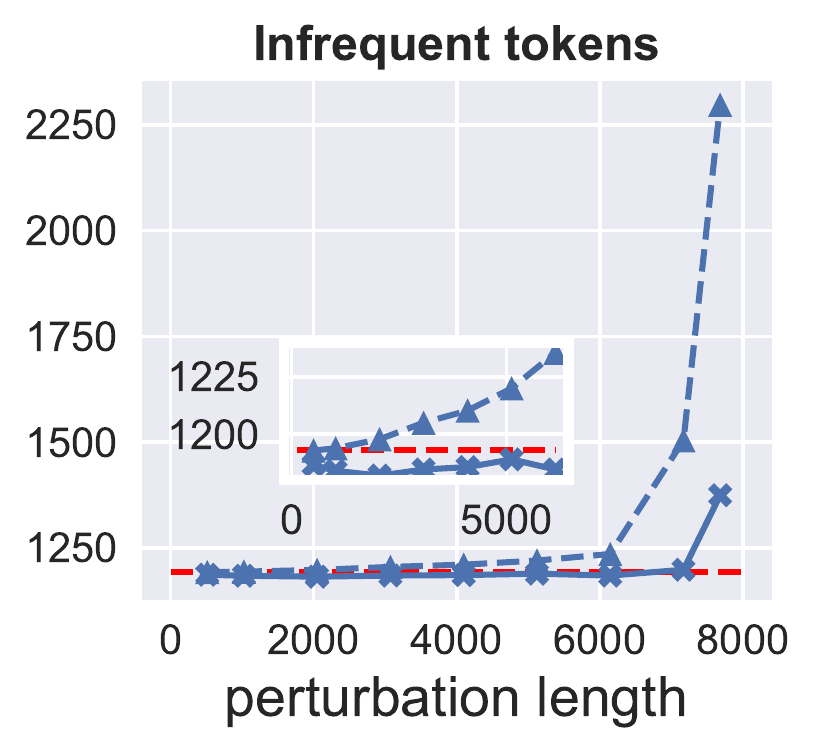}
    \end{minipage}
    \caption{
    While perturbing up to 6K prefix tokens has little impact on frequent tokens, it increases the perplexity of infrequent tokens.
    }
    \label{fig:perturb-overall-freq-vs-length}
\end{figure}

\begin{figure}
    \centering
    \begin{minipage}{0.225\textwidth}
    \includegraphics[width=\textwidth]{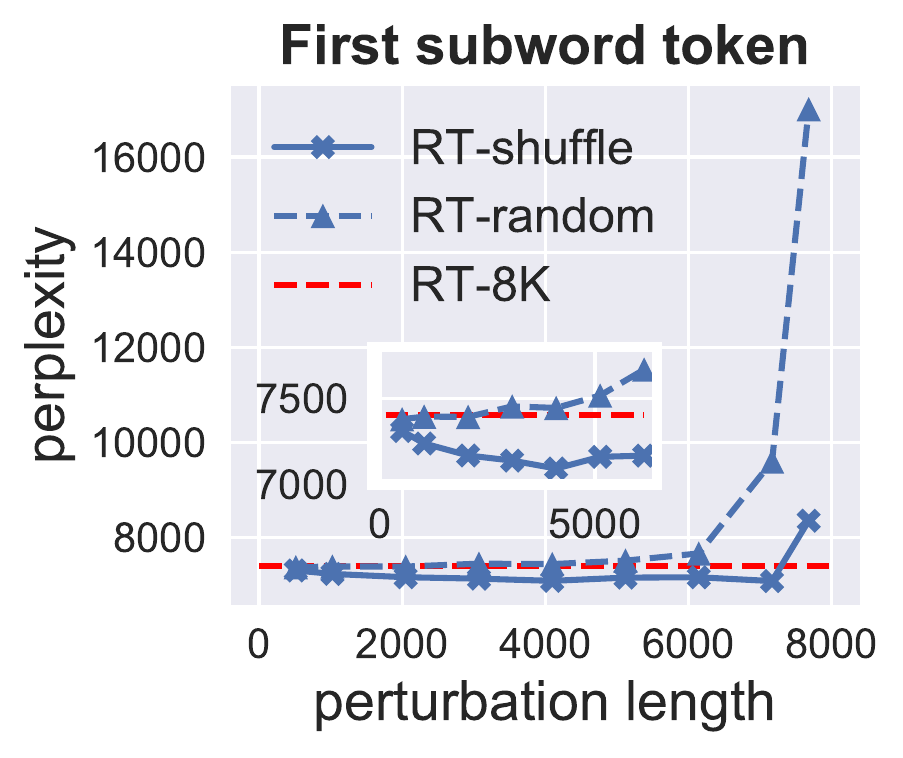}
    \end{minipage}
    \centering
    \begin{minipage}{0.20\textwidth}
    \includegraphics[width=\textwidth]{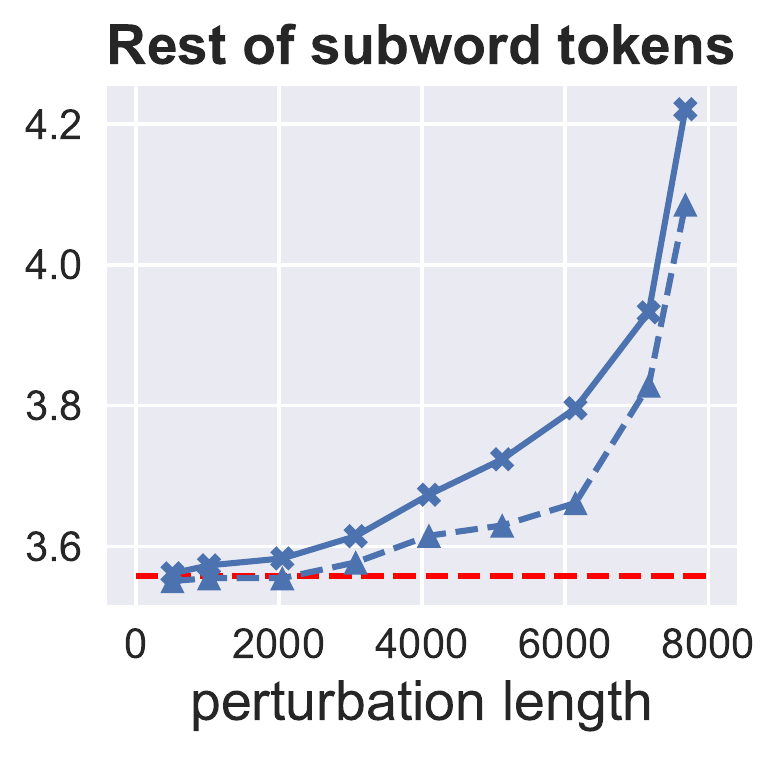}
    \end{minipage}
    \caption{
    Both shuffling and random replacement increase the perplexity of tokens inside subword clusters, with the former having more negative impact.
    % Perplexity of first token (\textbf{left}) and the rest tokens (\textbf{right}) in the subword cluster with perturbed distant prefix.
    }
    \label{fig:perturb-subwords-vs-length}
\end{figure}

\paragraph{Sequence-level perturbations further than 2K tokens from the target have minimal impact on perplexity:} We first apply sequence-level shuffling and random replacement to the distant context. Both operations have minimal impact on the perplexity of all tokens (Figure~\ref{fig:perturb-overall-vs-length}) as well as frequent/infrequent tokens (Figure~\ref{fig:perturb-overall-freq-vs-length}) provided at least 2K tokens are left unperturbed. However, these perturbations do have increasing impact as the perturbations come closer to the target, especially for infrequent tokens. Zooming in on the long-range context, we find that random replacement consistently results in higher perplexity than shuffling, but also that shuffling distant context actually achieves slightly \emph{lower} perplexity than when the model is given completely unperturbed prefixes. Overall, these results demonstrate that RT is insensitive to the word order of the long-range context.

\begin{figure}
    \centering
    \begin{minipage}{0.225\textwidth}
    \includegraphics[width=\textwidth]{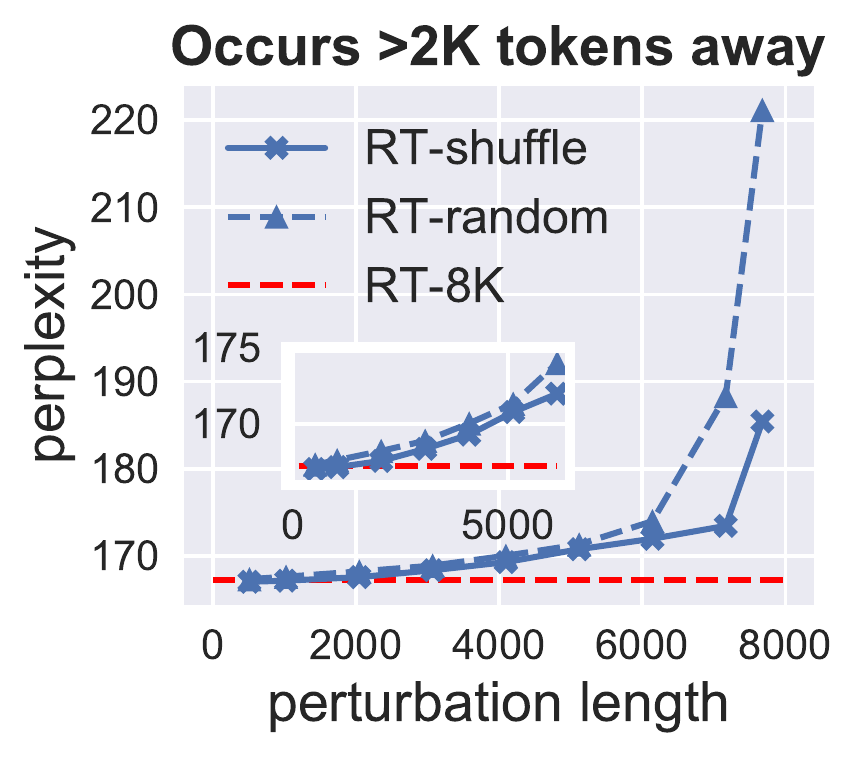}
    \end{minipage}
    \centering
    \begin{minipage}{0.215\textwidth}
    \includegraphics[width=\textwidth]{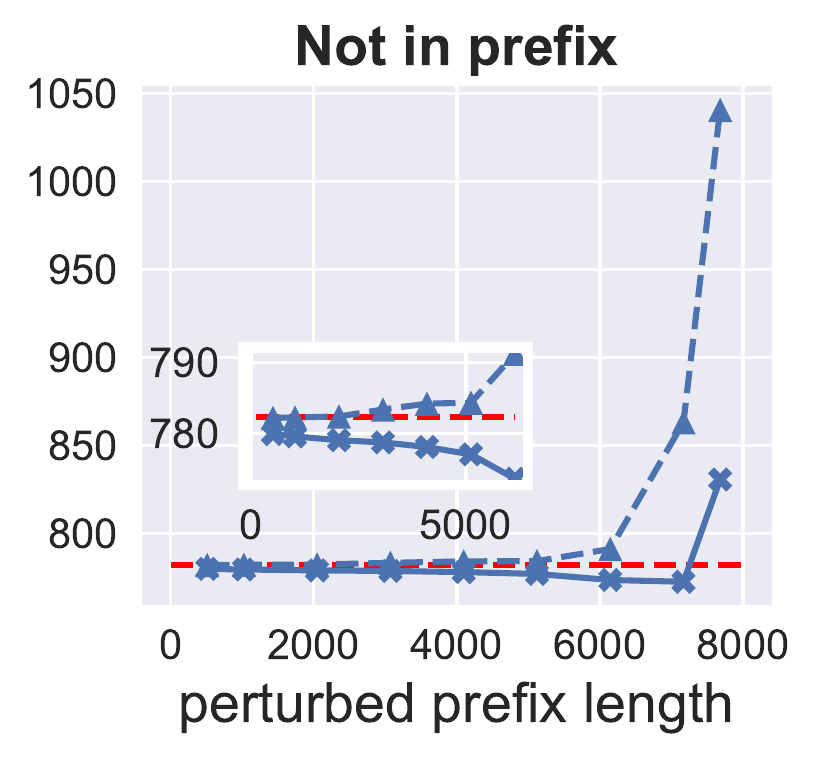}
    \end{minipage}
    \caption{
    Both perturbation operations increase the perplexity of target tokens whose last appearance in the prefix is more than 2K tokens away.
    % Perplexity of tokens that can only be found 2K tokens prior to target (\textbf{left}) and those never appear in the prefix (\textbf{right}) with perturbed distant prefix.
    }
    \label{fig:perturb-copy-vs-length}
\end{figure}

\paragraph{Tokens inside subword clusters and tokens repeated in the distant context depend on word order:} Similar to the analysis in Section~\ref{section:context-size}, the experiments above may hide impacts on small subsets of tokens, which motivates us to do a more fine-grained analysis.
We find that tokens inside subword clusters (Figure~\ref{fig:perturb-subwords-vs-length}) and those that can only be copied from long-range context (Figure~\ref{fig:perturb-copy-vs-length}) are sensitive to both the order and the content of the remote context. 
% Note that for random replacement, copied tokens are preserved while all of the surrounding context is replaced.
While random replacement is more harmful than shuffling for tokens that  can be copied in the distant context (172 shuffled vs 174 random replacement perplexity when perturbing 6K tokens),  shuffling is more detrimental for tokens inside subword clusters (3.8 vs 3.7 perplexity when perturbing 6K tokens).
% While random replacement has more impact on the latter, providing shuffled distant prefix is more harmful than presenting random content for tokens inside subword cluster.
% \micomment{put some more specific numbers in the writing e.g. how much is perplexity increase w/ shuffling}

% \paragraph{Word order of distant context matters less than content}  We find in the zoomed insets that shuffling the distant context can achieve slightly better perplexity for infrequent tokens (Figure~\ref{fig:perturb-overall-freq-vs-length} right), the begin token of subword cluster (Figure~\ref{fig:perturb-subwords-vs-length} left), and those never appear in the prefix (Figure~\ref{fig:perturb-copy-vs-length} right), however, all of them are sensitive to random replacement. This suggests the prediction of these tokens is based on fuzzy semantic field of the distant scope where order of tokens does not matter.

\begin{figure}
    \centering
    \begin{minipage}{0.235\textwidth}
    \includegraphics[width=\textwidth]{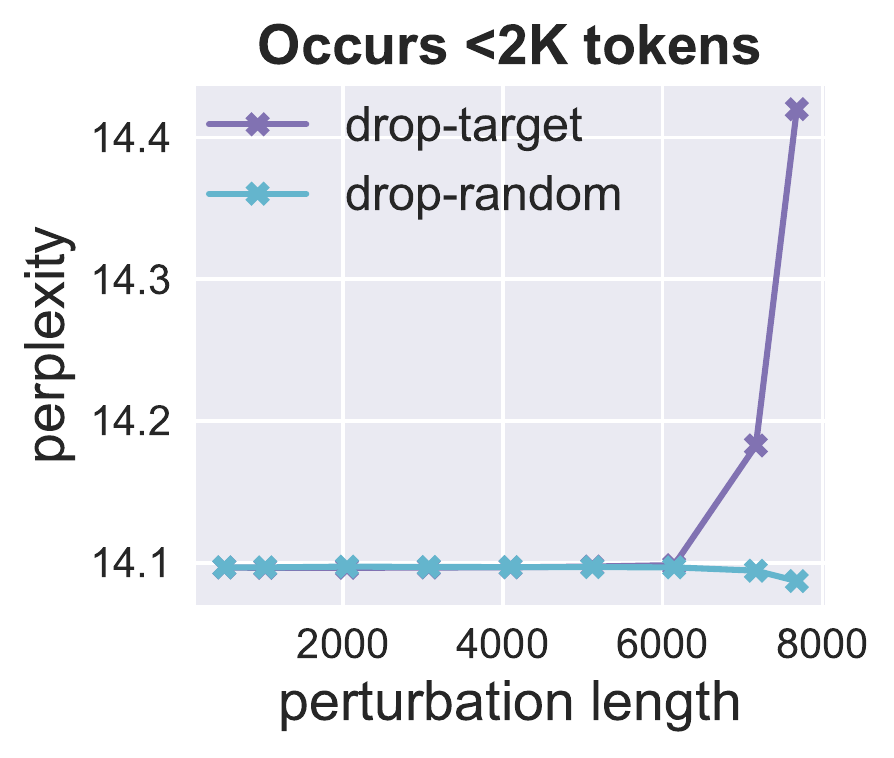}
    \end{minipage}
    \centering
    \begin{minipage}{0.215\textwidth}
    \includegraphics[width=\textwidth]{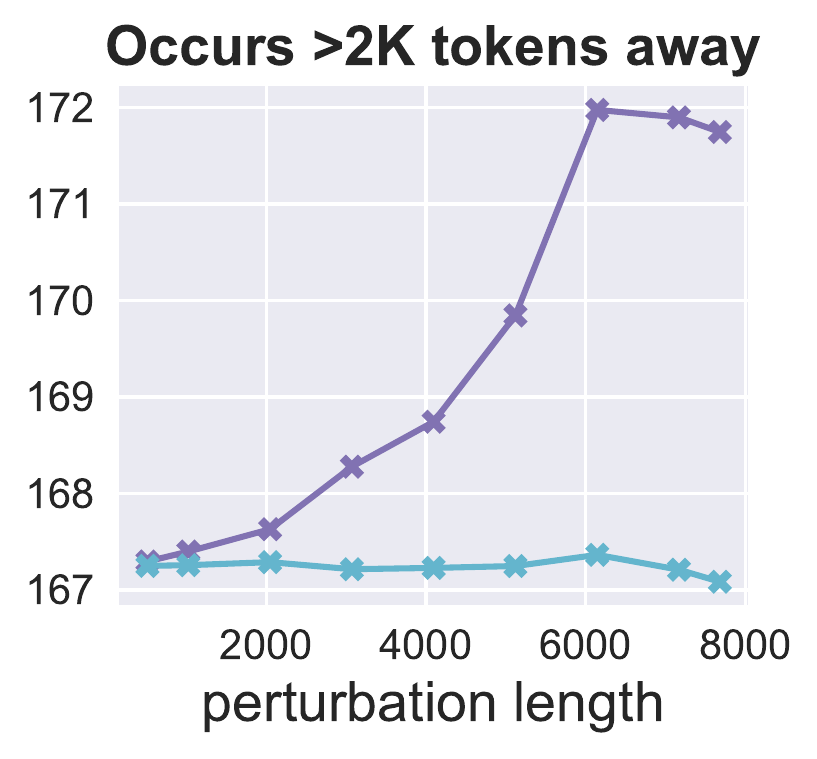}
    \end{minipage}
    \caption{
    Perplexity of target tokens whose last occurrence in the prefix is within previous 2K tokens (\textbf{left}) or more than 2K tokens away (\textbf{right}), when dropping either these target tokens or random tokens in the perturbed range. The curve on the right indicates RT memorizes token identity in the distant prefix to some extent.
    }
    \label{fig:perturb-target-vs-length}
\end{figure}

\paragraph{Routing Transformer encodes token identity in the long-range context:} 
While the previous perturbations affected entire contiguous blocks of the prefix, we move now to more targeted perturbations. An interesting question to ask given the observation that RT perplexity decreases on copied tokens as sequence length increases (\S~\ref{section:context-size}) is how much that perplexity decrease depends on word order and surrounding content. In response, we drop tokens in the distant context whose next appearance is in the target sequence. As a control experiment, we drop the same number of random tokens for each perturbation length.
% \micomment{what does within the same range mean?}
%middle

As shown in the right plot of Figure~\ref{fig:perturb-target-vs-length}, dropping the previous long-range occurrences of target tokens increases the perplexity of those target tokens, which shows that RT indeed memorizes token identity in the long-range context to some extent.
%left
The left plot shows that dropping long-range duplicate tokens does not affect tokens that also occur within the local context (i.e., the prior 2K tokens). The flat curve before 6K indicates the model relies only on the most recent occurrences for prediction.
%right
% The perturbed target tokens in long-range context may constitute necessary information for those do not appear in the prefix, thus shows a slightly higher value than unperturbed baseline. Dropping random tokens when the unperturbed context size is small actually decreases perplexity. This is likely because there are fewer constraints in the input, so model assigns higher probability to previously unrelated tokens.

\section{Sequence-level analysis}

% teacher forcing skew the prediction
% token-level eval may hide info
% summary of each section

All of the previous experiments have focused on \emph{token-level perplexity}, which is the standard way in which LMs are evaluated. However, the prefixes in these evaluations consist solely of ground-truth text, mirroring the ``teacher-forcing'' setup that LMs are trained with. When these models are deployed practically to generate text, they have to rely on their previous predictions instead of ground-truth text, and several prior works have noted different behavior in this setting~\cite{wang-sennrich-2020-exposure,Holtzman2020The,Welleck2020Neural}. In this section, we shift from token-level tasks to analyzing RT and LT performance on \emph{sequence-level} tasks. In particular, we first look at how well the models can memorize an exact sequence in the distant context, as opposed to a single token as we did previously. Next, we examine the models' ability to identify which of six 128-token suffixes follows a given prefix, which examines their behavior outside the standard teacher-forced setting. 
% Language models are usually trained with teacher forcing, i.e., the ground-truth tokens in previous steps are accepted as input for predicting future tokens. Always providing the correct recent tokens during training may skew the target token prediction to not learning dependencies in the long-range context. The previous analyses are all on the token-level under teacher forcing scheme, which conflicts the generation purpose of language models at inference time. 

\begin{figure}
    \centering
    \begin{minipage}{0.24\textwidth}
        \includegraphics[width=\textwidth]{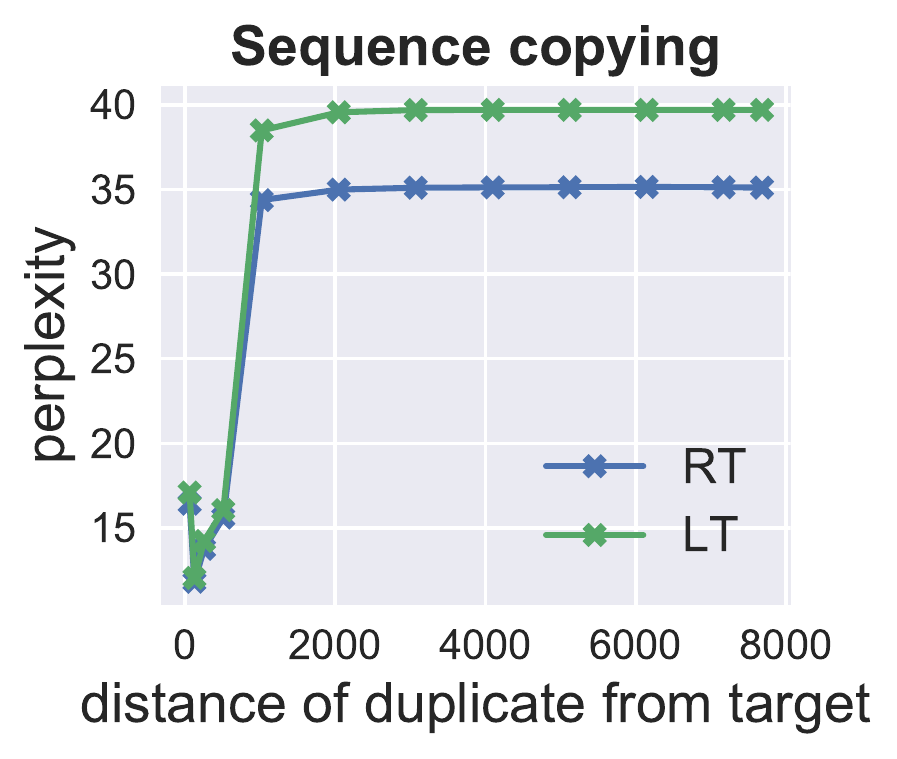}
    \end{minipage}
    \begin{minipage}{0.225\textwidth}
        \includegraphics[width=\textwidth]{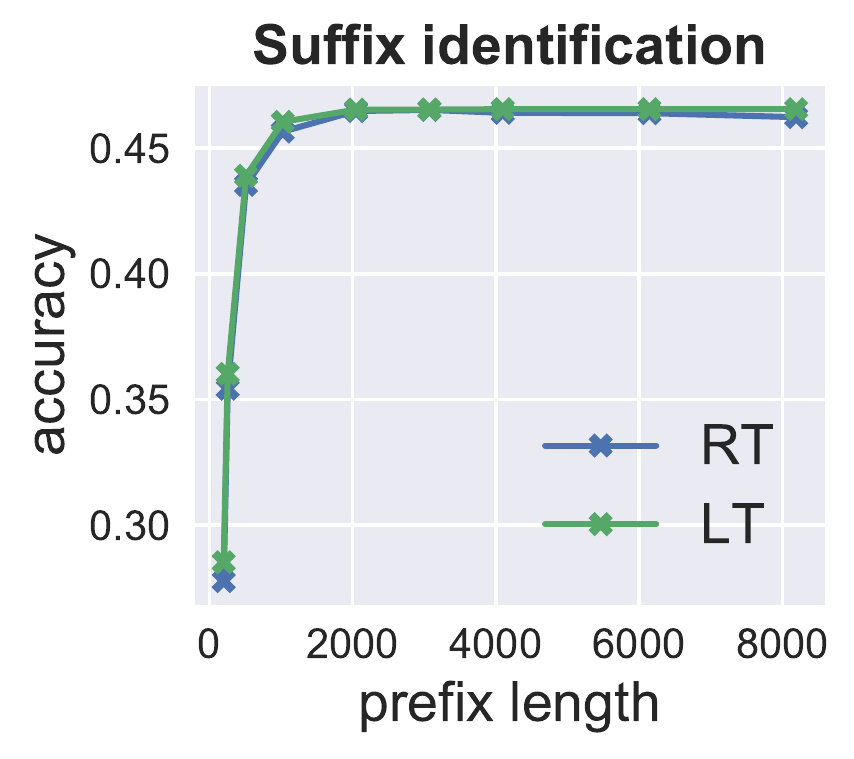}
    \end{minipage}
    \caption{\textbf{Left:} 
    % Perplexity of target sequence with duplicate placing at different distance. 
    Both models assign low perplexity if a duplicate sequence appears within previous 512 tokens. \textbf{Right:}
    % Suffix identification accuracy when evaluated with different prefix length. 
    Both models have almost identical performance on our suffix identification task. Adding context beyond 2K tokens does not improve performance of either sequence-level prediction task.}
    \label{fig:copy-past-tgt}
\end{figure}

\paragraph{Sequence-level copying:} \label{section:copy-analysis}
As a sequence-level analogue to the token-copying analysis in the previous section, we examine both RT and LT's ability to memorize a sequence that occurs in the distant context.
To test this ability, we copy the target sequence and paste it into different positions of the prefix. The left plot in Figure~\ref{fig:copy-past-tgt}  shows that both models give a very low perplexity to the target sequence if its duplicate appears within the previous 512 tokens. However, \textbf{both models lose their ability to take advantage of the copied sequence if it occurs more than 2K tokens away}. This confirms our previous discovery that sequence order is in general not encoded in the long-range context.

\begin{table}[]
    % \footnotesize
    \scalebox{0.85}{
    \begin{tabular}{p{0.25\linewidth}p{0.75\linewidth}}
    \toprule
     \textbf{Prefix}    & $\dots$
    %  If you die to-morrow I lose my money.
     If the \textcolor{blue}{doctor}'s prophecy is correct$\dots$ \textcolor{gray}{($\sim$700 tokens)} $\dots$
    %  have his opinion about the whole matter."
     "How far is it to his place?""Oh, a mile at least.We can have a cab.""A mile? \\\hline
     \textbf{Gold suffix}:    &  Then we shall see if there is any truth in what that swab of a \textcolor{blue}{doctor} said $\dots$ \\\hline
     \textbf{Negative 1}: &  If I can see Mr.Elberry to-day we may let you have a cheque $\dots$\\
     \textbf{Negative 2}: & It was not until he had signed and sent it off that the full significance of all$\dots$\\
     \textbf{Negative 3}: & He hurried in, fearing that she might have taken some turn for the worse$\dots$\\
     \textbf{Negative 4}: & look!!!"Her voice had fallen suddenly to a quivering whisper and she was$\dots$\\
     \textbf{Negative 5}: & Again the Admiral burst out cheering."There remains, therefore$\dots$\\
     \bottomrule
    \end{tabular}
    }
    \caption{An example of suffix identification task, the full version of this example is included in Appendix~\ref{section:appendix-suffix-identification}.}
    \label{tab:suffix-identification}
\end{table}

\paragraph{Suffix identification:} 
To move beyond token-level experiments, we adopt a similar setting as the multiple choice task in SWAG~\citep{zellers-etal-2018-swag}.
Specifically, a prefix is paired with the ground-truth next 128 tokens (or \emph{suffix}) as well as five randomly sampled sequences of length 128 that come from the same book and do not occur in the prefix or gold suffix. We constrain the prefix to end at a full stop and each candidate suffix to start from a new sentence so that the difference in perplexity is not due to ungrammaticality. An example is shown in Table~\ref{tab:suffix-identification}.
We construct 7K examples and compute the accuracy of both models at correctly choosing the correct suffix. The model makes a correct prediction when the gold suffix has lower perplexity than all other suffixes.
As shown in the right plot of Figure~\ref{fig:copy-past-tgt}, increasing prefix length beyond 2K does not improve suffix identification accuracy. Surprisingly, the LT and RT model have almost identical (and poor) performance on this task.\footnote{While evaluating on the newly released LT checkpoint, the performance of LT is slightly worse, but the trend is similar. Adding context beyond 2K tokens does not keep improving the suffix identification accuracy. We direct reader to Appendix~\ref{section:appendix-new-lt} for more details.} While RT is a significantly better LM in terms of token-level perplexity, it does not appear to be superior in terms of using long-range context to improve sequence prediction. Overall, both models often predict obviously wrong negative suffixes: the full version of Table~\ref{tab:suffix-identification} together with RT's perplexity score for each suffix is included in Appendix~\ref{section:appendix-suffix-identification}. 

Combined with our previous token-level analysis, we conclude that the distant context helps a subset of tokens in superficial ways; however, distant context is currently not helpful for sequence-level prediction tasks.

\section{Related work}

% long-range language models
Our work examines recent advances in efficient Transformer variants~\cite{sukhbaatar-etal-2019-adaptive,Kitaev2020Reformer,choromanski2021rethinking, tay2021synthesizer,katharopoulos20,wang2020linformer, wu2020lite} that accept longer sequences than prior approaches. Longer effective context size is often achieved by sparse attention~\cite{child2019generating}, recurrence~\cite{dai-etal-2019-transformer}, and cached memory~\cite{weston2015memory,Rae2020Compressive}. Our work is also related to methods that incorporate long context~\cite{wang-cho-2016-larger} as well as document-level tasks that inherently require modeling long-range context~\cite{zhang-etal-2018-improving, hofstatter2020local, zhang-etal-2020-long}.

% analysis on long-range context
This work is also similar to other analysis of language models, especially for long-range context.~\citet{khandelwal-etal-2018-sharp} analyze the usage of long-term context of smaller LSTM LMs. ~\citet{10.1145/3188745.3188954} prove long-term context is not needed for HMM LM due to teacher forcing. ~\citet{rae-razavi-2020-transformers} conduct an analysis exclusively for the Transformer-XL~\cite{dai-etal-2019-transformer} model. ~\citet{Rae2020Compressive} show that Compressive Transformer improves the performance of infrequent tokens. Our work also relates to that of~\citet{lai-etal-2020-context}, who investigate the impact of context for pretrained masked LMs.  More recently, ~\citet{press2020shortformer} also observe negligible benefits of long-term context; we step further in this direction by exploring  larger models with more fine-grained analysis. 
\section{Conclusion}

We perform a fine-grained analysis of the impact of long-range context to both token- and sequence-level improvements on two long-range Transformer language models, using the PG-19 dataset as a testbed. Our results suggest these models rarely take advantage of the long-term context, and when they do it is mostly in superficial ways (e.g, by copying rare tokens from far away). With the proliferation of research in increasing the input size of Transformer LMs, we hope that our research will lead to more meaningful progress on integrating discourse information into these models. 
\section*{Ethical concerns}

\paragraph{Misuse of language models} The two large language models we evaluated in this work share common ethical concerns with works on language models and language generation. These pre-trained LMs can be used maliciously to generate unfaithful, hallucinated, and biased output. Our reported results do not include any kind of generation.

\paragraph{Energy costs} We conduct all our analysis experiments on RTX8000 GPUs. Although our work does not include training large language models, the energy costs of evaluating large pre-trained LMs, such as the Routing Transformer, should not be ignored. Each example of 8K tokens long takes around 1.3s $\sim$ 1.4s to run one forward pass with the RT model. We hope our analysis can shed light on more efficient and effective method to encode long-term context.

\section*{Acknowledgements}

We are grateful to Aurko Roy for releasing the code and checkpoints and for discussing the Routing Transformer results with us. We thank Nader Akoury, Brendan O'Connor and the rest of UMass NLP group for the great advice on the draft of this paper. We thank the anonymous reviewers for their thoughtful comments on the paper. This project was partially funded by a grant from Intuit AI and also by award IIS-1955567 from the National Science Foundation (NSF).

\bibliography{anthology,custom}
\bibliographystyle{acl_natbib}

\appendix

\begin{figure*}
    \centering
    \includegraphics[width=0.75\textwidth]{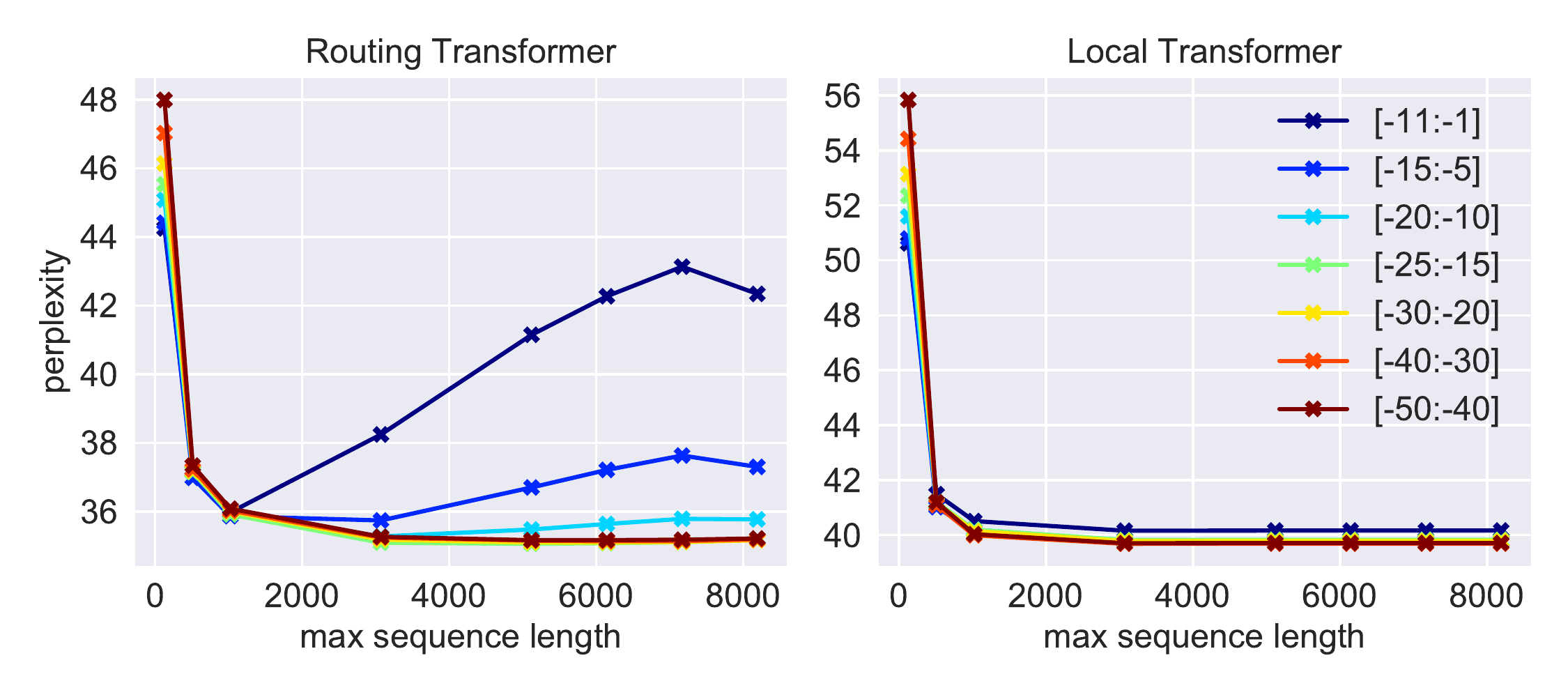}
    \caption{Perplexity of target chunk of 10 tokens long near the end of sequence. The legend indicates how far away the evaluated chunk is from the end of sequence (e.g.,  [-15:-5] means evaluating the last 15 tokens to the last 5 tokens). The left plot shows the clustering mechanism in the Routing Transformer assigns higher perplexity to around the last 15 tokens in a sequence. Due to this artifacts, in our analysis, we avoid tokens within this range to make comparable comparison with the Local Transformer. In the main text, our analysis are conducted over the last 50 to last 40 tokens, which are not affected by this artifact. }
    \label{fig:end-seq-issue}
\end{figure*}

\newpage 
\section{Routing Transformer and end-sequence degradation}

In our analysis, instead of picking the last 10 tokens in a sequence, we chose the last 50 to last 40 tokens due to an artifact introduced by the clustering heads in the RT model. We find that in general the last 20 tokens in a sequence tend to have increasing perplexity as we evaluate on longer and longer sequence lengths. As shown in Figure~\ref{fig:end-seq-issue}, this phenomenon is only native to the RT model and disappears when the clustering heads are replaced with local attentions. Therefore, to make the RT and the LT comparable, we select the tokens from the range that is not affected by the end-sequence issue. Although it's not the last 10 tokens, this short target chunk is still located near the end of a sequence, preceded by enough long context.

\begin{figure*}
    \centering
    \includegraphics[width=0.3\textwidth]{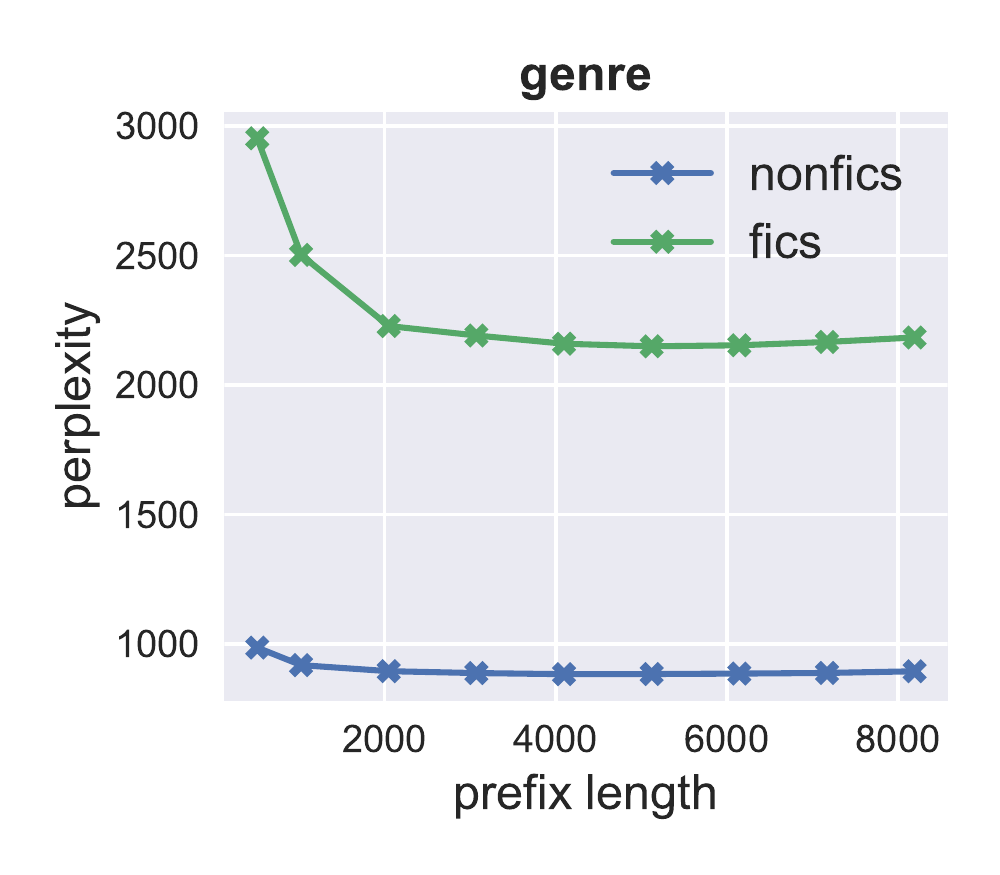} \includegraphics[width=0.3\textwidth]{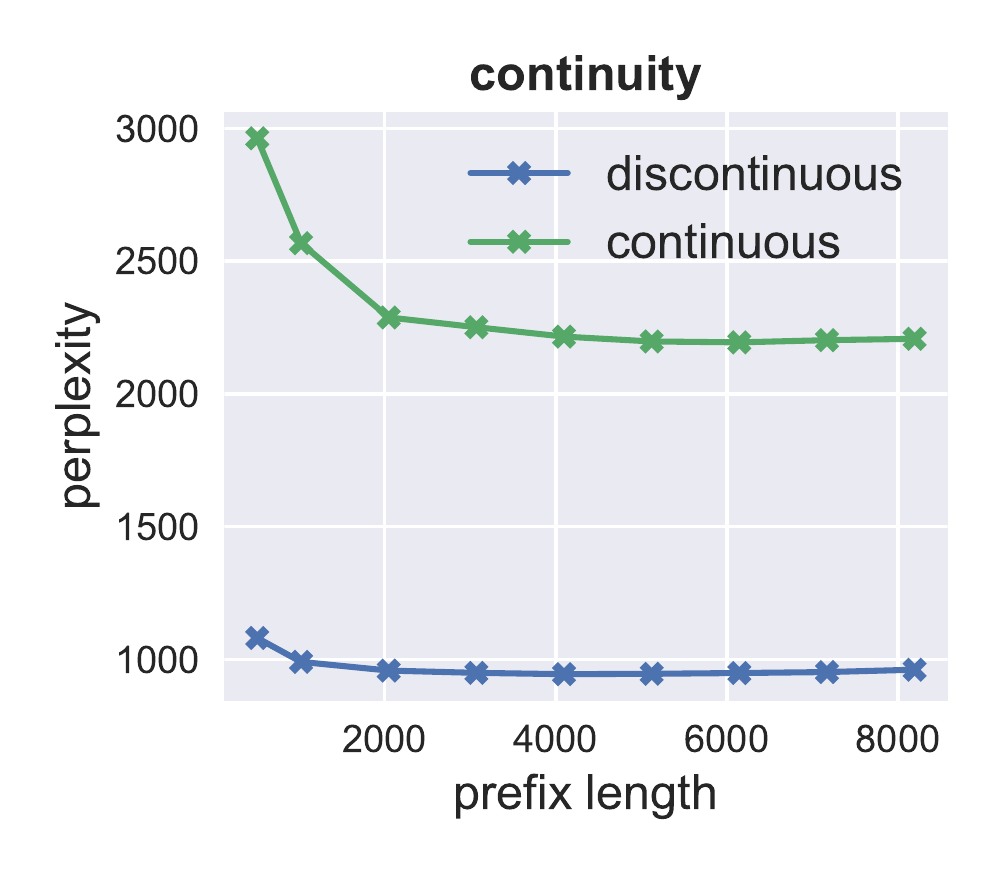} \includegraphics[width=0.3\textwidth]{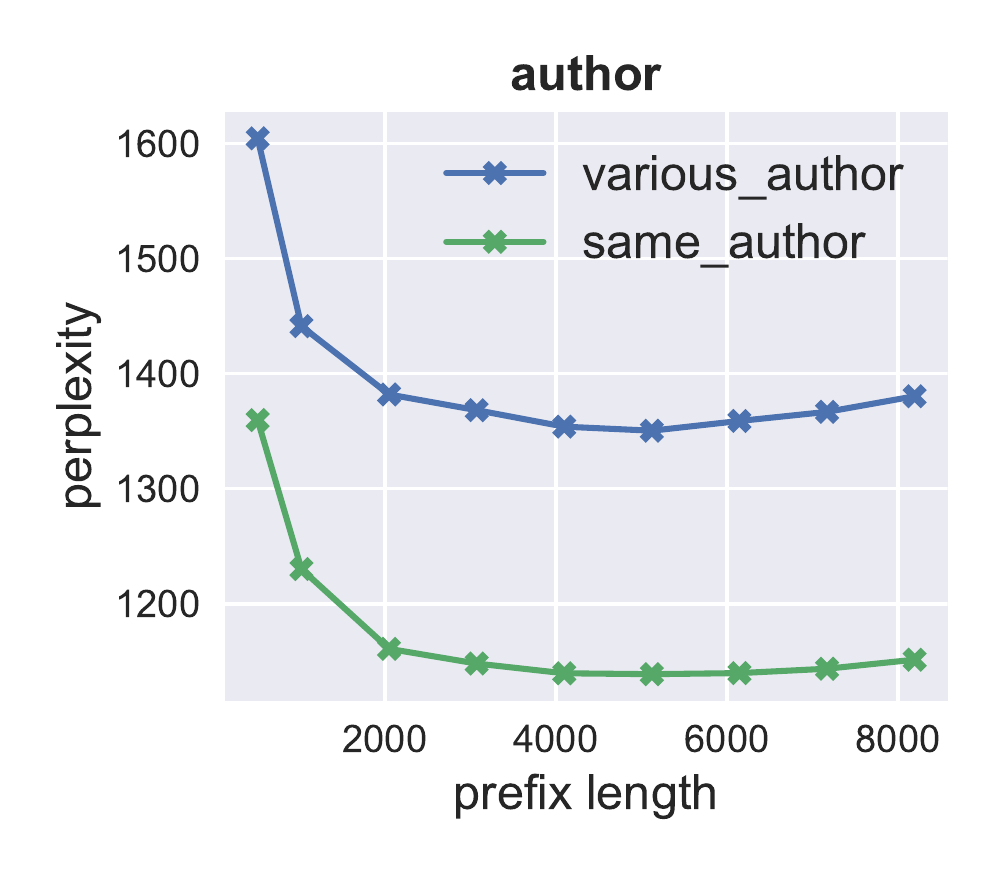}
    \caption{Perplexity of infrequent target tokens, broken down by genre (\textbf{left}), continuity (\textbf{middle}), and authorship (\textbf{right}). Perplexity of infrequent tokens in fictional, continuous and single-authored books decreases as the context length increases to around 5K. On the other hand, the rest types of books rely on more local context while predicting the infrequent tokens.}
    \label{fig:freq-genre-book-types}
\end{figure*}

\begin{figure*}
    \centering
    \includegraphics[width=0.3\textwidth]{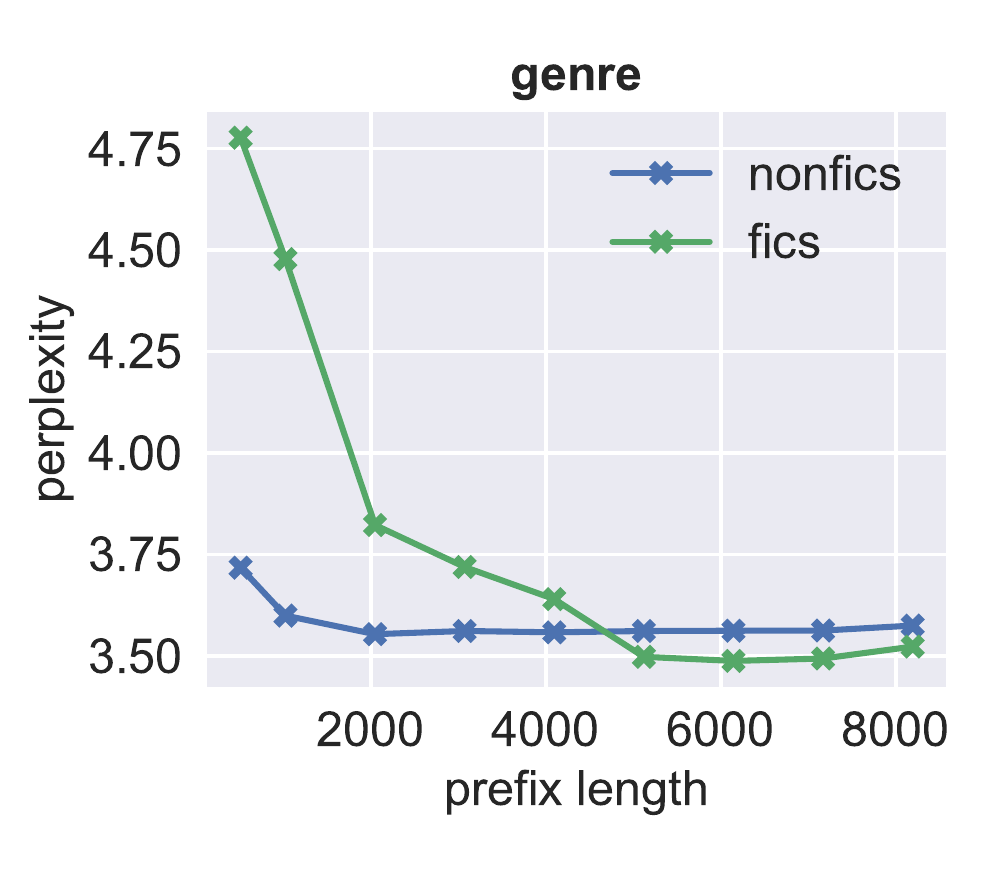} \includegraphics[width=0.3\textwidth]{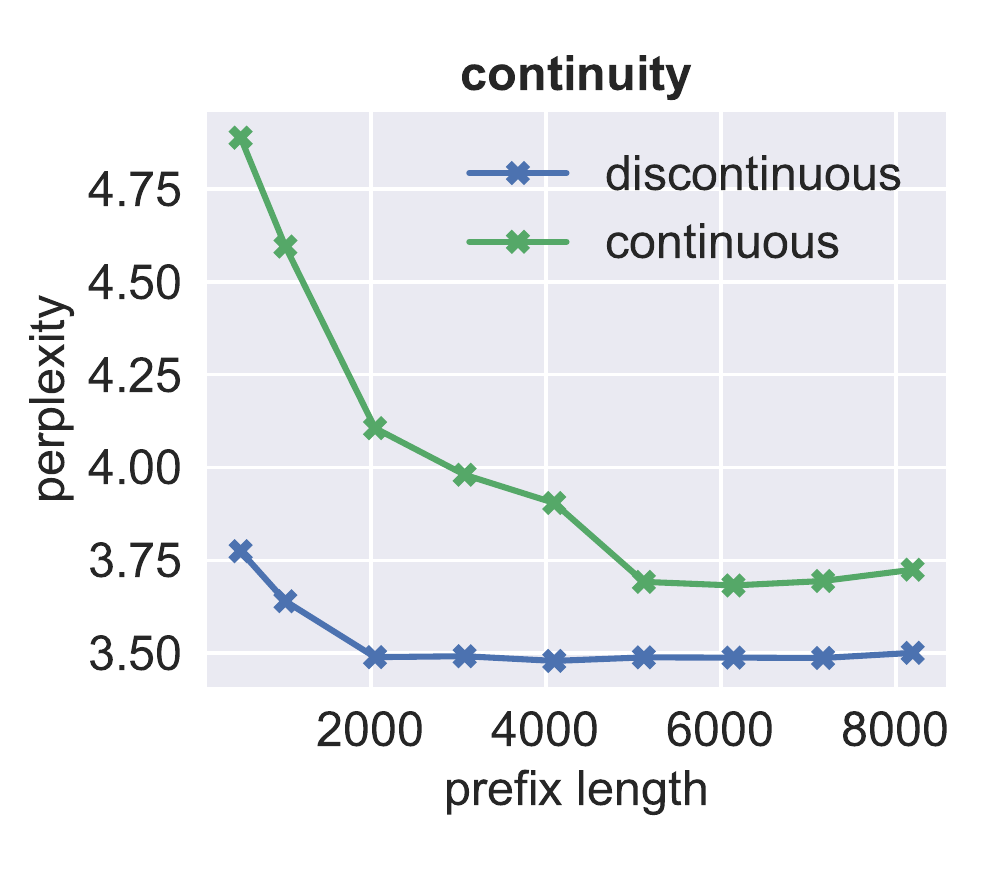} \includegraphics[width=0.3\textwidth]{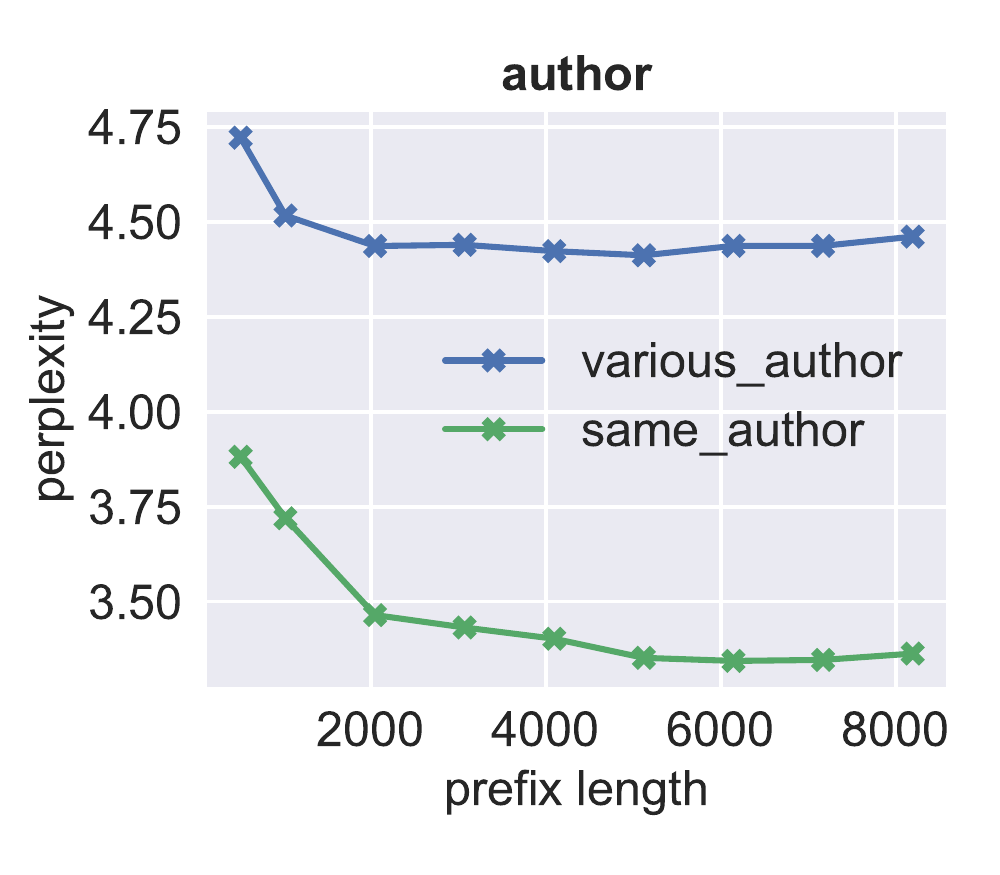}
    \caption{Perplexity of target tokens inside subword clusters (i.e., excluding the first subword in each cluster), broken down by genre (\textbf{left}), continuity (\textbf{middle}), and authorship (\textbf{right}). Perplexity of these tokens in fictional and continuous books improves as the length increases up to around 4K, whereas nonfictional and discontinuous books are not taking any advantage of long-range context at all.   }
    \label{fig:subword-genre-book-types}
\end{figure*}

\section{Effect of longer context} \label{section:appendix-book-type}
In section~\ref{section:context-size} we discussed that books that are fictional and continuous benefit more from the long-range context. We also annotated the validation set by the authorship (i.e., whether a book is written by single author or various authors). Out of 49 books, 11 are written by various authors, 10 of which are non-fictions. Due to this overlap, we only show results of fic/non-fic in the main text. 

\begin{figure*}
    \centering
     \includegraphics[width=0.9\textwidth]{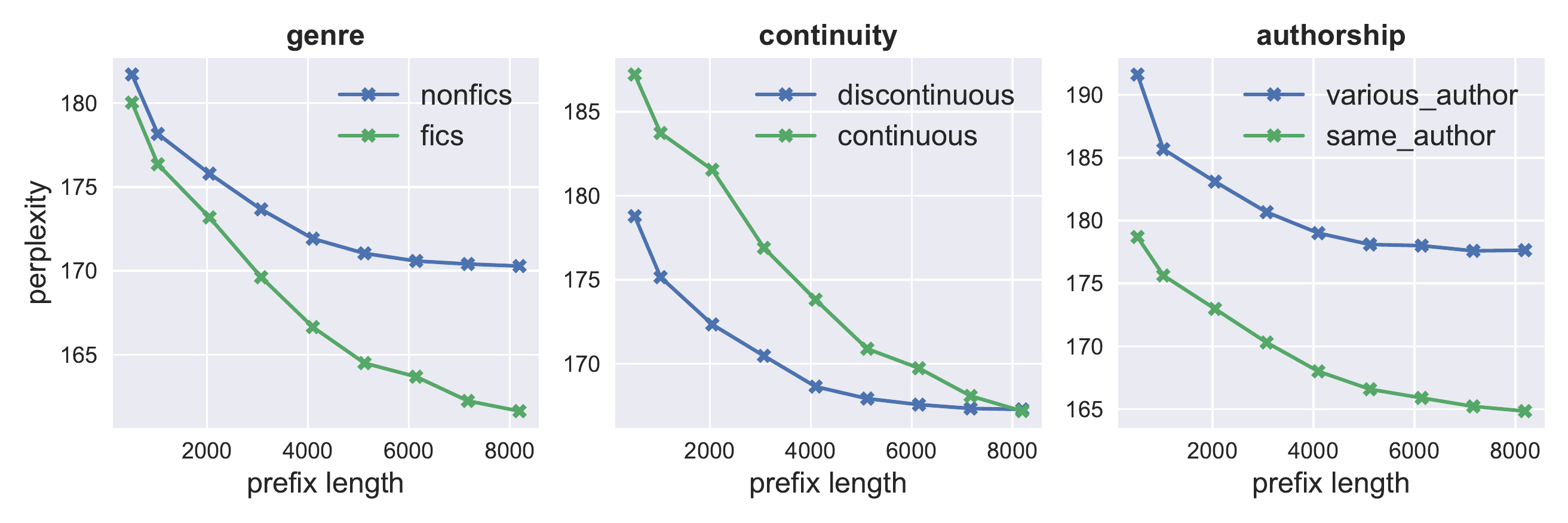}
    \caption{Perplexity of target tokens that can only be found in the distant context, when evaluated with the Routing Transformer on subset of PG-19 validation set, broken down by genre (\textbf{left}), continuity (\textbf{middle}), and authorship (\textbf{right}). Fictional, continuous, and single-authored books continue to have improved perplexity for this type of tokens as the prefix length increases up to 8K. The single-authored books also contain discontinuous books which require less modeling of long-range context. The decreasing curve indicates the model might have acquired author specific token statistics from incorporating longer context.}
    \label{fig:distant-by-book-types-all}
\end{figure*}

In this section , we also further break down all targets to the three types of tokens we examined in section~\ref{section:context-size}, and display the results by book types. Perplexity of infrequent tokens (Figure~\ref{fig:freq-genre-book-types}), tokens inside subword clusters (Figure~\ref{fig:subword-genre-book-types}), and tokens whose last occurrence is more than 2K tokens away (Figure~\ref{fig:distant-by-book-types-all}). In general, for the small set of tokens whose perplexity keep decreasing as adding in more context, the major source of improvements are from the continuous and fictional books. 

\begin{figure}[th]
    \centering
    \includegraphics[width=0.35\textwidth]{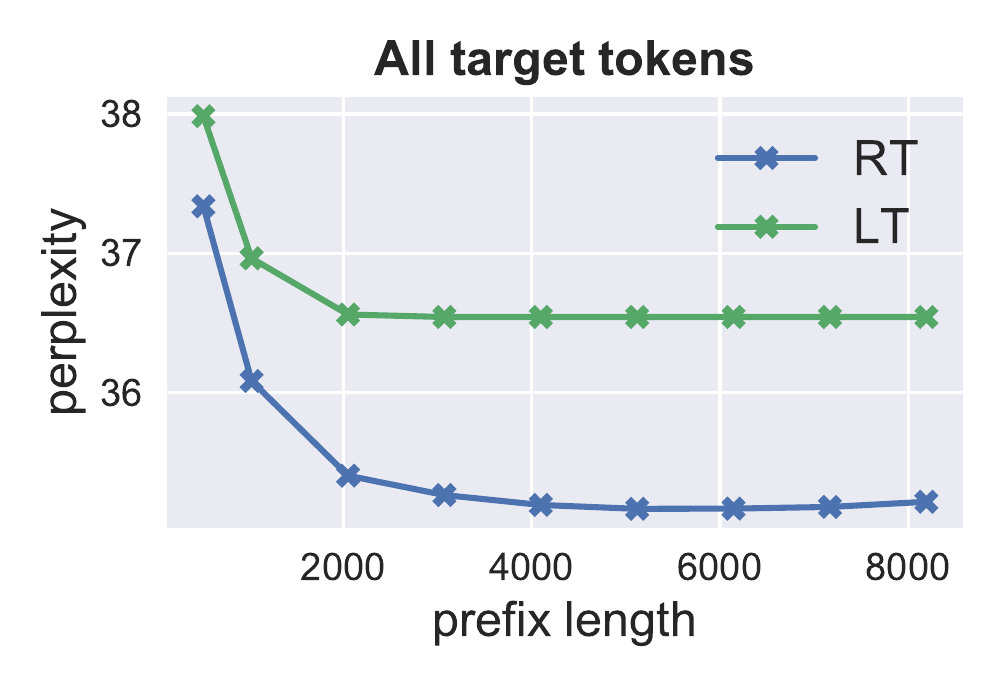}
    \caption{The perplexity of all target tokens plateaus after 2K prefix tokens for both Routing Transformer and Local Transformer. The LT model is the one released in 2021 summer.}
    \label{fig:perplexity-by-length-new-lt}
\end{figure}

\begin{figure}[th]
    \centering
    \includegraphics[width=0.35\textwidth]{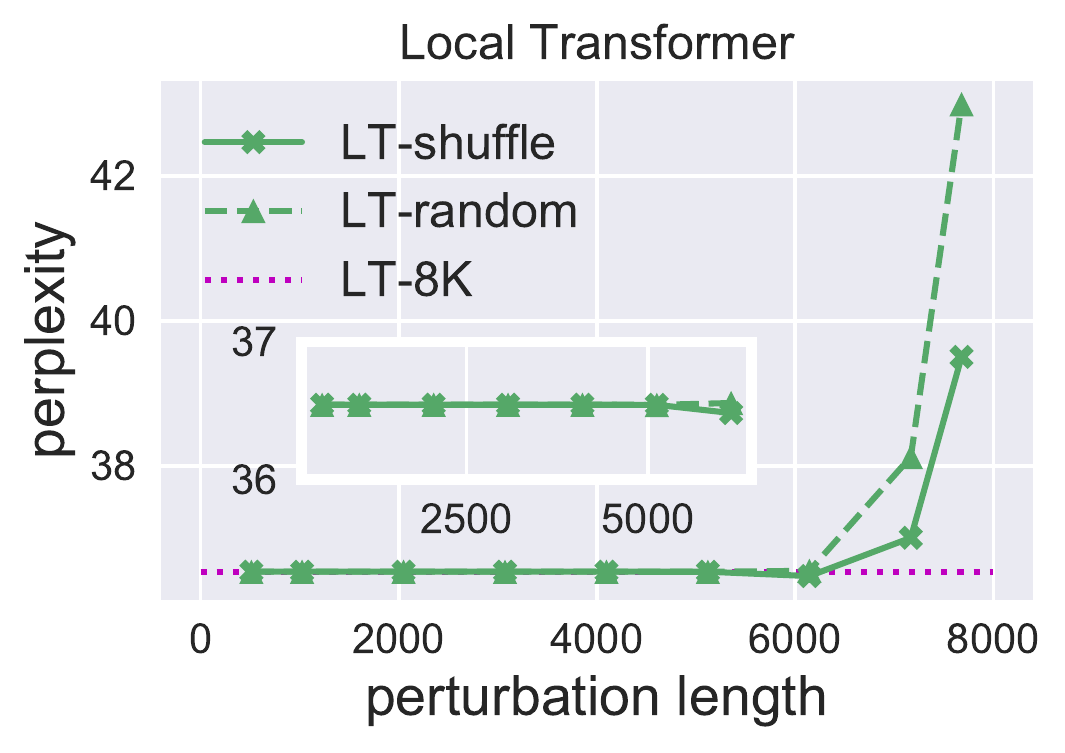}
    \caption{Perplexity of all target tokens when evaluated with Local Transformer released in 2021 summer. }
    \label{fig:perturb-overall-LT-new}
\end{figure}

\begin{figure}[th]
    \centering
    \includegraphics[width=0.35\textwidth]{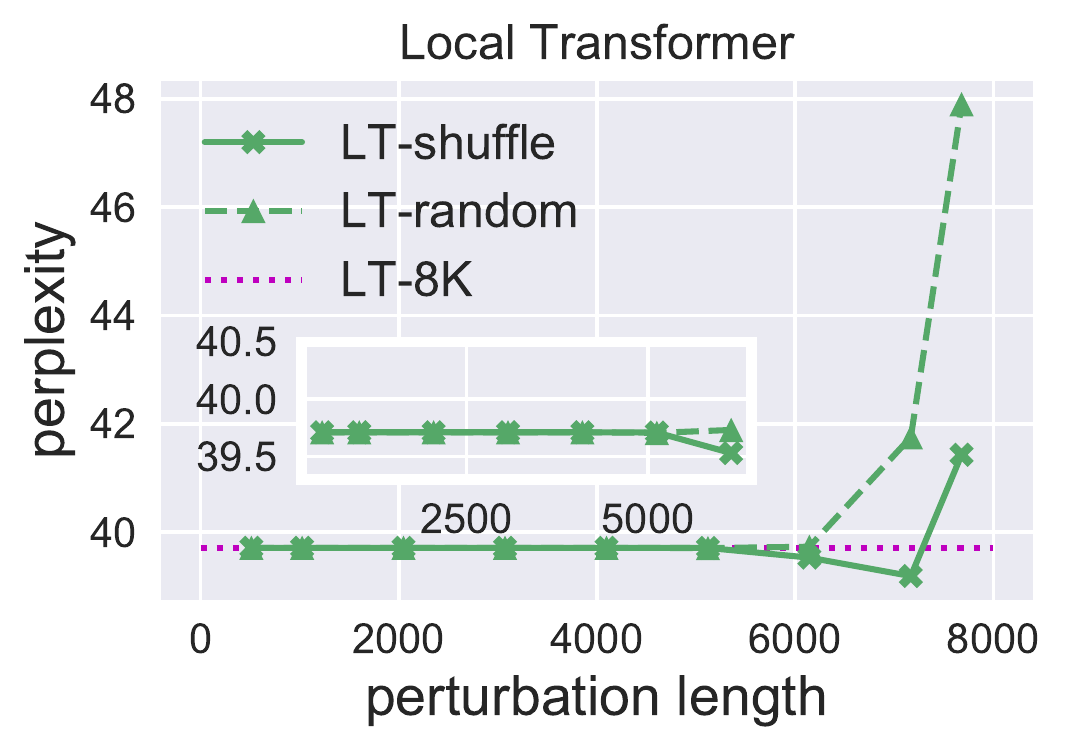}
    \caption{Perplexity of all target tokens when evaluated with Local Transformer derived from the RT checkpoint. }
    \label{fig:perturb-overall-LT}
\end{figure}

\begin{table}[th]
    \centering
    \begin{tabular}{c|ccc}
    \hline
    & infreq & in-subword & distant \\\hline
    infreq & 1. & 0.09 & 0.08  \\
    in-subword & 0.36 & 1. & 0.1 \\
    distant & 0.07 & 0.02 & 1. \\\hline
	
    \end{tabular}
    \caption{Ratio of overlapped target tokens among different categorizations. \textbf{infreq} are infrequent tokens, \textbf{in-subword} are tokens within a subword cluster (i.e., excluding the first word in a subword cluster), \textbf{distant} are tokens only appear in the distant context (more than 2K away). Row 1 column 3 the number 0.08 indicates around 8\% of the infrequent tokens are those that can only be found in the distant context.}
    \label{tab:overlap-ratio}
\end{table}

\begin{figure}
    \centering
    \includegraphics[width=0.35\textwidth]{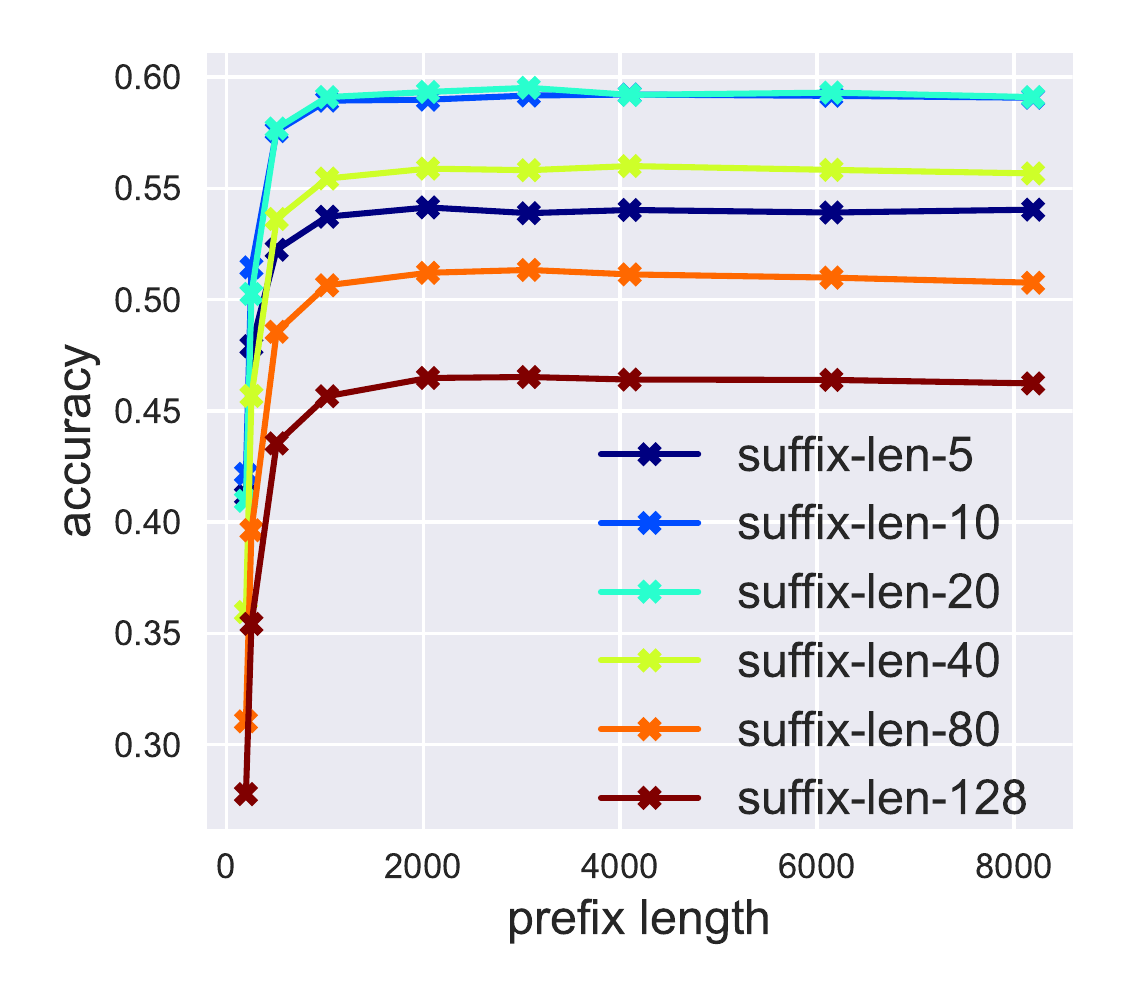}
    \caption{Suffix identification accuracy when evaluated with different prefix length.}
    \label{fig:suffix-by-seq-len-all-suffix-len}
\end{figure}

\section{Token overlaps} \label{section:token-overlaps}
We have shown in section~\ref{section:context-size} that infrequent tokens, tokens inside subword clusters, and tokens that can only be copied from distant context benefit from context longer than 2K tokens.
It is possible that these improvements come from the same set of tokens shared across these three types of tokens. To verify if there are significant overlaps among those three types of tokens, we compute the overlapped ratio in Table \ref{tab:overlap-ratio}. Except for in-subword and infrequent tokens, the overlapped ratios are all below 0.1.

\section{Perturbation} \label{section:appendix-context-perturb}
Perplexity with perturbed distant prefix when evaluated with Local Transformer is shown in Figure~\ref{fig:perturb-overall-LT}. Perplexity hardly changes when perturbing up to around 6K tokens. Because LT doesn't properly use long-range context, we only present the results of Routing Transformer in the main text. 

\section{Suffix Identification}\label{section:appendix-suffix-identification}
In the main text, we present the suffix identification results with 128-token long suffix. Here, we provide results when evaluate the accuracy with suffix of different length.
Interestingly, the accuracy of distinguishing a gold suffix first increases with the suffix length, reaching the best when the suffix length is around 10 to 20, then decreases as the suffix becomes longer. This is likely because the sequence becomes more probable as incorporating more local context (part of the suffix).

In Table~\ref{tab:suffix-identification-example-prefix} and Table~\ref{tab:tab:suffix-identification-example-negatives}, we present a complete example of the suffix identification task. The prefix contains 1024 tokens, and each of the suffixes contains 128 tokens. Lower perplexity of obviously wrong suffix (e.g. negative 1,3,4) indicates current RT model is not properly taking advantage of distant context to make sequence-level predictions.

\section{Local Transformer checkpoint results} \label{section:appendix-new-lt}

In the main text, we analyzed both the RT checkpoint and an LT model derived from the same RT checkpoint by replacing the clustering heads with local attention heads. After the submission deadline and before the camera ready deadline, the author of the Routing Transformer released a new LT checkpoint, which has 24 layers in total\footnote{Both the RT model and the \emph{former} LT model have 22 layers.}. To make sure the behavior of our former LT is the same with an LT trained from scratch, we also conducted all our analysis again on this new checkpoint. Overall, we find the new LT checkpoint performs slightly better on token-level tasks but is inferior to the previous LT checkpoint on suffix identification, a sequence-level task. Since the trends are the same, no conclusion needs to be changed. 

\paragraph{The Effect of Longer Context} In Figure~\ref{fig:perplexity-by-length-new-lt} are the perplexities in aggregate over all target tokens of both the RT and the new LT checkpoints. The target tokens are the same ones we used in the main text. The new LT has better perplexity than the one presented in the main text (36.5 vs. 40 as the prefix length extends to 8K). 

\paragraph{Perturbation of long-range context} Similar to the former LT (Figure~\ref{fig:perturb-overall-LT}), the newly released LT checkpoint(Figure~\ref{fig:perturb-overall-LT-new}) is not sensitive at all to the perturbation further than local 2K tokens. Both models are impacted by local random replacement more than shuffling, however, the new LT checkpoint has overall better perplexity than the RT-derived LT.

\begin{figure}
    \centering
    \begin{minipage}{0.24\textwidth}
        \includegraphics[width=\textwidth]{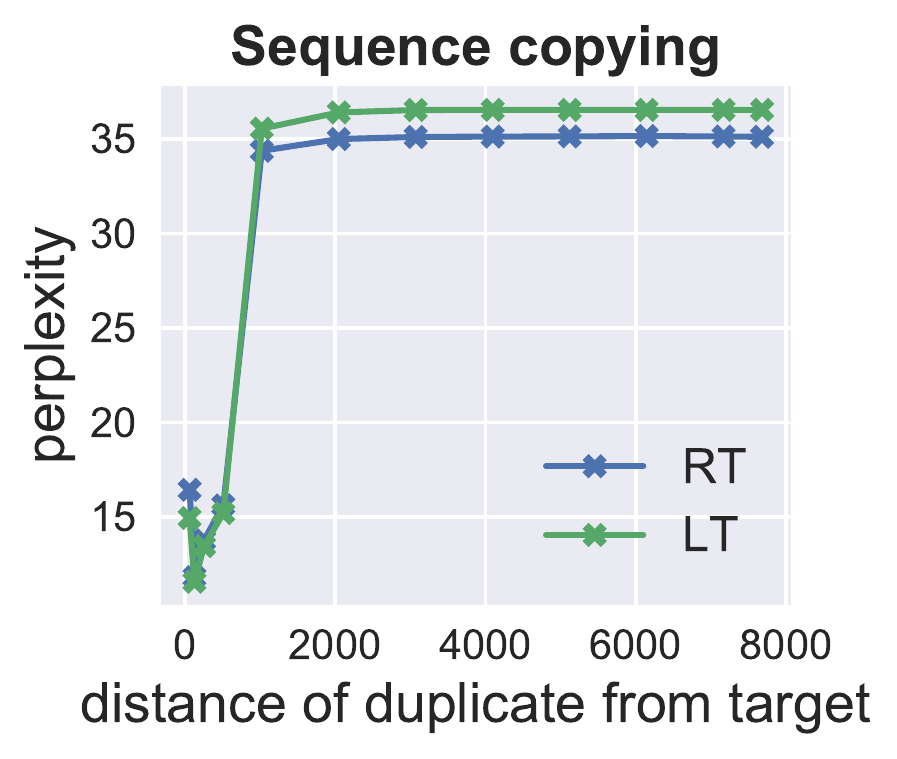}
    \end{minipage}
    \begin{minipage}{0.225\textwidth}
        \includegraphics[width=\textwidth]{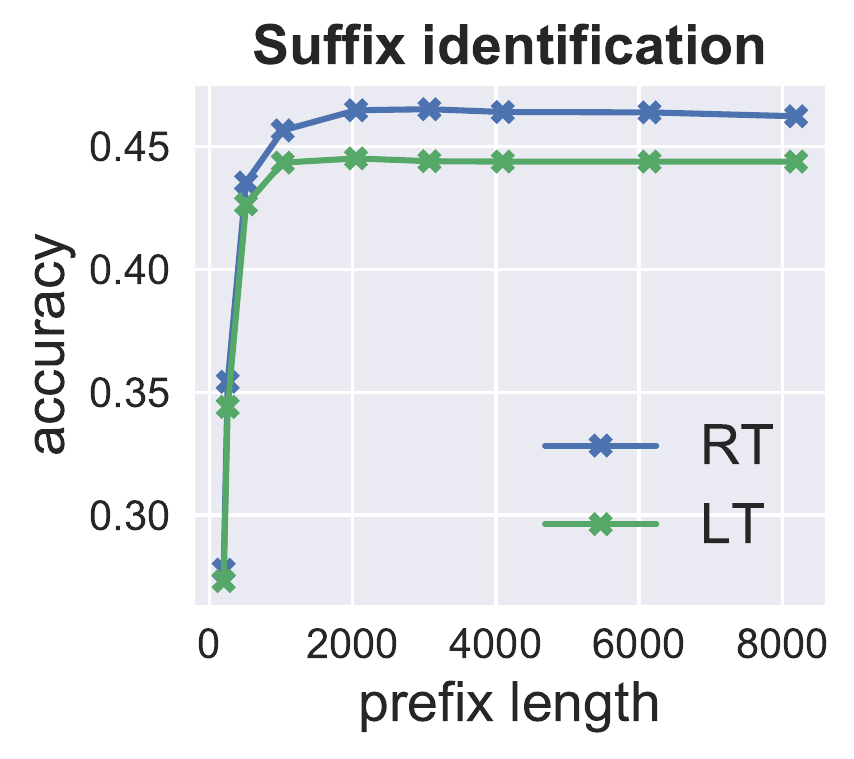}
    \end{minipage}
    \caption{\textbf{Left:} 
    Both models assign low perplexity if a duplicate sequence appears within previous 512 tokens. \textbf{Right:}
    Adding context beyond 2K tokens does not improve performance of suffix identification. Moreover, the recently released LT performs worse than the one derived from the RT checkpoint.}
    \label{fig:copy-past-tgt-new}
\end{figure}

\paragraph{Sequence-level tasks} Figure~\ref{fig:copy-past-tgt-new} shows the performance of both RT and the released LT checkpoint on sequence-level tasks. Compared to the results in the main text, the new LT checkpoint performs better at sequence-copying task, however, the trend remains the same -- the order of tokens beyond 2K tokens is not properly encoded. On the other hand, the new LT checkpoint is slightly worse in suffix identification while the former RT-derived LT has almost identical performance as the RT. This implies even though the clustering heads are removed from the previous LT, useful information is preserved by the local attention heads.

Overall, we find the released LT checkpoint has better token-level performance but performs worse on suffix identification. Each plot shares the same trend with its corresponding one in the main text, thus no conclusion needs to be modified. 

\begin{table*}[]
    \centering
   \scalebox{0.8}{
    \begin{tabular}{p{0.1\linewidth}p{0.83\linewidth}}
    \toprule
     Prefix    & Not a bit, sir.Out with it!I have faced death too often to flinch from it now, though I saw it as near me as you are.""Well, well, we must go by averages of course.Shall we say two years?I should think that you have a full two years before you.""In two years your pension would bring you in L1,600.Now I will do my very best for you, Admiral!I will advance you L2,000, and you can make over to me your pension for your life.It is pure speculation on my part.If you die to-morrow I lose my money.If the doctor's prophecy is correct I shall still be out of pocket.If you live a little longer, then I may see my money again.It is the very best I can do for you.""Then you wish to buy my pension?""Yes, for two thousand down.""And if I live for twenty years?""Oh, in that case of course my speculation would be more successful.But you have heard the doctor's opinion.""Would you advance the money instantly?""You should have a thousand at once.The other thousand I should expect you to take in furniture.""In furniture?""Yes, Admiral.We shall do you a beautiful houseful at that sum.It is the custom of my clients to take half in furniture."The Admiral sat in dire perplexity.He had come out to get money, and to go back without any, to be powerless to help when his boy needed every shilling to save him from disaster, that would be very bitter to him.On the other hand, it was so much that he surrendered, and so little that he received.Little, and yet something.Would it not be better than going back empty-handed?He saw the yellow backed chequebook upon the table.The moneylender opened it and dipped his pen into the ink."Shall I fill it up?"said he."I think, Admiral," remarked Westmacott, "that we had better have a little walk and some luncheon before we settle this matter.""Oh, we may as well do it at once.It would be absurd to postpone it now," Metaxa spoke with some heat, and his eyes glinted angrily from between his narrow lids at the imperturbable Charles.The Admiral was simple in money matters, but he had seen much of men and had learned to read them.He saw that venomous glance, and saw too that intense eagerness was peeping out from beneath the careless air which the agent had assumed."You're quite right, Westmacott," said he."We'll have a little walk before we settle it.""But I may not be here this afternoon.""Then we must choose another day.""But why not settle it now?""Because I prefer not," said the Admiral shortly."Very well.But remember that my offer is only for to-day.It is off unless you take it at once.""Let it be off, then.""There's my fee," cried the doctor."How much?""A guinea."The Admiral threw a pound and a shilling upon the table."Come, Westmacott," said he, and they walked together from the room."I don't like it," said Charles, when they found themselves in the street once more; "I don't profess to be a very sharp chap, but this is a trifle too thin.What did he want to go out and speak to the doctor for?And how very convenient this tale of a weak heart was!I believe they are a couple of rogues, and in league with each other.""A shark and a pilot fish," said the Admiral."I'll tell you what I propose, sir.There's a lawyer named McAdam who does my aunt's business.He is a very honest fellow, and lives at the other side of Poultry.We'll go over to him together and have his opinion about the whole matter.""How far is it to his place?""Oh, a mile at least.We can have a cab.""A mile? \\\hline
     Gold suffix (ppl=74.4)    &  Then we shall see if there is any truth in what that swab of a doctor said.Come, my boy, and clap on all sail, and see who can stay the longest."Then the sober denizens of the heart of business London saw a singular sight as they returned from their luncheons.Down the roadway, dodging among cabs and carts, ran a weather-stained elderly man, with wide flapping black hat, and homely suit of tweeds.With elbows braced back, hands clenched near his armpits, and chest protruded, he scudded along, while close at his heels lumbered a large-limbed, heavy, yellow mustached young man, who seemed to feel the exercise a good deal more than his senior.On they dashed, helter-skelter, until they pulled up panting at the office where the lawyer of the \\
     \bottomrule
    \end{tabular}
   }
    \caption{Suffix identification example, extracted from the book \emph{Beyond the City} by Arthur Conan Doyle. For each suffix, we show the perplexity of the suffix evaluated with the Routing Transformer. RT assigns a lot lower perplexity to examples (e.g. negative 1,3,4) that are obviously wrong given the prefix.}
    \label{tab:suffix-identification-example-prefix}
\end{table*}

\begin{table*}[]
    \centering
    \scalebox{0.8}{
    \begin{tabular}{p{0.1\linewidth}p{0.83\linewidth}}
    \toprule
     Negative 1 (ppl=34.6): &  If I can see Mr.Elberry to-day we may let you have a cheque to-morrow.Try another pinch.No?Well, good-bye.I am very happy to have been of service."Mr.McAdam bowed them out, for he was a very busy man, and they found themselves in the street once more with lighter hearts than when they had left it."Well, Westmacott, I am sure I am very much obliged to you," said the Admiral."You have stood by me when I was the better for a little help, for I'm clean out of my soundings among these city sharks.But I've something to do now which is more in my own line, and I need not trouble you any more.""Oh, it is no trouble.I have nothing\\\hline
     Negative 2 (ppl=71.69): & It was not until he had signed and sent it off that the full significance of all that he had done broke upon him.He had sacrificed everything.His pension was gone.He had nothing save only what he could earn.But the stout old heart never quailed.He waited eagerly for a letter from the Saint Lawrence Shipping Company, and in the meanwhile he gave his landlord a quarter's notice.Hundred pound a year houses would in future be a luxury which he could not aspire to.A small lodging in some inexpensive part of London must be the substitute for his breezy Norwood villa.So be it, then!Better that a thousand fold than that his name should be associated with failure and disgrace.On that morning Harold Denver was to meet the creditors of the firm, and to explain the situation to them.It was a hateful task\\\hline
     Negative 3 (ppl=33.54): & He hurried in, fearing that she might have taken some turn for the worse, but he was reassured to find her sitting up in her bed, with Clara and Ida Walker in attendance upon her.She had removed the handkerchief, and had put on a little cap with pink ribbons, and a maroon dressing-jacket, daintily fulled at the neck and sleeves."My dear friend," said she as he entered, "I wish to make a last few remarks to you.No, no," she continued, laughing, as she saw a look of dismay upon his face."I shall not dream of dying for at least another thirty years.A woman should be ashamed to die before she is seventy.I wish, Clara, that you would ask your father to step up.And you, Ida, just pass me my cigarettes, and\\\hline
     Negative 4 (ppl=34.92): & look!!!"Her voice had fallen suddenly to a quivering whisper and she was pointing to the Westmacotts' house.Her sister gave a gasp of horror, and stood with a clutch at Monica's arm, staring in the same direction.There was a light in the front room, a slight, wavering light such as would be given by a small candle or taper.The blind was down, but the light shone dimly through.Outside in the garden, with his figure outlined against the luminous square, there stood a man, his back to the road, his two hands upon the window ledge, and his body rather bent as though he were trying to peep in past the blind.So absolutely still and motionless was he that in spite of the moon they might well have overlooked him were it not for that tell-tale light behind.\\\hline
     Negative 5 (ppl=47.66): & Again the Admiral burst out cheering."There remains, therefore, about L3,200 which has to be found within ten days.No man shall lose by me.I gave them my word in the room that if I worked my soul out of my body every one of them should be paid.I shall not spend a penny upon myself until it is done.But some of them can't wait.They are poor men themselves, and must have their money.They have issued a warrant for Pearson's arrest.But they think that he has got away to the States.""These men shall have their money," said the Admiral."Dad!""Yes, my boy, you don't know the resources of the family.One never does know until one tries.What have you yourself now?""I have about a\\
     \bottomrule
    \end{tabular}
    }
    \caption{Suffix identification example, extracted from the book \emph{Beyond the City} by Arthur Conan Doyle. For each suffix, we show the perplexity of the suffix evaluated with the Routing Transformer. RT assigns a lot lower perplexity to examples (e.g. negative 1,3,4) that are obviously wrong given the prefix.}
    \label{tab:tab:suffix-identification-example-negatives}
\end{table*}
\label{sec:appendix}

\end{document}